\documentclass[journal]{IEEEtran}
\usepackage{cite}
\usepackage{url}
\usepackage{times}
\usepackage{amsmath,graphicx}
\usepackage{epstopdf}
\usepackage{xcolor}
\usepackage{amssymb}
\usepackage{array}
\usepackage{booktabs}
\usepackage{multirow}
\usepackage{graphicx}
\usepackage{color}
\usepackage{float}
\usepackage{stfloats}
\usepackage[misc]{ifsym}
\usepackage{stmaryrd}
\usepackage{grffile}
\usepackage[caption=false,font=footnotesize]{subfig}
\usepackage{makecell}
\usepackage{float}
\usepackage{textcomp}
\usepackage{threeparttable}
\usepackage{enumitem}
\usepackage{setspace}
\usepackage[pagebackref=true,breaklinks=true,linkcolor=red,anchorcolor=blue, citecolor=green,colorlinks,bookmarks=false,backref=false]{hyperref}
\usepackage{tabularx}
\usepackage{upgreek}

\definecolor{LightBlue}{rgb}{0.88,0.92,0.95}
\definecolor{Orange}{rgb}{1,0.75,0}
\definecolor{DarkBlue}{rgb}{0,0,1}

\begin{document}
	
\title{Infrared Small Target Detection in Satellite Videos: A New Dataset and A Novel Recurrent Feature Refinement Framework}%

\author{Xinyi~Ying, Li~Liu, Zaipin Lin, Yangsi Shi, Yingqian Wang, Ruojing Li, Xu Cao, Boyang Li, Shilin Zhou, Wei An
	
	\thanks{X.~Ying, L.~Liu, Z.~Lin, Y.~Shi, Y.~Wang, X.~Cao, R.~Li, S.~Zhou, W.~An are with the College of Electronic Science and Technology, National University of Defense Technology, P. R. China. Zaiping~Lin and Wei~An are the corresponding authors. Emails: yingxinyi18@nudt.edu.cn.
	This work was partially supported by the National Key Research and Development Program of China No. 2021YFB3100800, the National Natural Science Foundation of China under Grant 62376283, the Science and Technology Innovation Program of Hunan Province under Grant 2021RC3069, the China Postdoctoral Science Foundation under Grant Number GZB20230982, 2023M744321, and the Independent Innovation Science Fund Project of National University of Defense Technology under Grant 22-ZZCX-042. 
	}}

\markboth{Submitted to IEEE Transactions on Geoscience and Remote Sensing}%
{Shell \MakeLowercase{\textit{et al.}}: Bare Demo of IEEEtran.cls for IEEE Journals}

\maketitle

\begin{abstract}
Multi-frame infrared small target (MIRST) detection in satellite videos has been a long-standing, fundamental yet challenging task for decades, and the challenges can be summarized as follows: First, extremely small target size, highly complex clutters \& noises, various satellite motions result in limited feature representation, high false alarms, and difficult motion analyses. In addition, existing methods are primarily designed for static or slightly adjusted perspectives captured by short-distance platforms, which cannot generalize well to complex background motion in satellite videos. Second, the lack of large-scale publicly available MIRST dataset in satellite videos greatly hinders the algorithm development. To address the aforementioned challenges, in this paper, we first build a large-scale dataset for MIRST detection in satellite videos (namely IRSatVideo-LEO), and then develop a recurrent feature refinement (RFR) framework as the baseline method for satellite motion estimation and compensation. Specifically, IRSatVideo-LEO is a semi-simulated dataset with synthesized satellite motion, target appearance, trajectory and intensity, which can provide a standard toolbox for satellite video generation and a reliable evaluation platform to facilitate the algorithm development. For the baseline method, RFR is proposed to be equipped with existing powerful CNN-based methods for long-term temporal dependency exploitation and integrated motion compensation \& MIRST detection. Specifically, a pyramid deformable alignment (PDA) module is proposed to achieve effective feature alignment, and a temporal-spatial-frequency modulation (TSFM) module is proposed to achieve efficient feature aggregation and enhancement. Extensive experiments have been conducted to demonstrate the effectiveness and superiority of our scheme. The comparative results show that ResUNet equipped with RFR outperforms the state-of-the-art MIRST detection methods. Dataset and code are released at \url{https://github.com/XinyiYing/RFR}.
\end{abstract}

\begin{IEEEkeywords}
	Infrared small target detection, satellite videos, recurrent network, deformable convolution.
\end{IEEEkeywords}

\section{Introduction}\label{sec-intro}
\IEEEPARstart{I}nfrared satellite surveillance is significant in diverse scenarios such as traffic monitoring \cite{DeepPrior,HiEUM,p-norm}, maritime rescue \cite{Hou2024,SRSTT} and early warning systems \cite{NQSA,MoCoPnet}. Despite its valuable applications, infrared satellite videos are always with strict access restrictions, resulting in data scarcity of open-source datasets. In addition, due to the technical limitations of satellite sensors, infrared satellite videos always exhibit low spatial \& radiometric resolution and limited imaging quality.

Multi-frame infrared small target detection in satellite videos (MIRST-SatVideo) is a long-standing, fundamental yet challenging task in satellite surveillance systems, and has consecutively received considerable attention. Specifically, MIRST detection aims at localizing a scarce of candidate target pixels from image sequences captured by diverse earth-orbiting satellites, such as low earth-orbiting (LEO) satellites of 400-2K km, middle earth-orbiting (MEO) satellites of 2K-36K km, and geostationary earth-orbiting (GEO) satellites of 36K km. Compared to IRST detection in aerial-based and land-based imaging systems (mostly no more than 10 km), IRST detection in satellite videos remains more challenging, and has several unique characteristics. 

\begin{itemize}
	\item \textbf{Extremely small size of targets:} Due to remote sensing imaging system and optical diffraction effect, targets always appear as small points or diffraction spots, which lack geometry appearances such as contour, shape and texture.
	\item \textbf{Highly complex clutters and noises:} Satellite videos always contain various complex background clutters (\textit{e.g.,} earth background clutter, stellar clutter and cloud clutter) and heavy sensor noise (\textit{e.g.,} random thermal noise and non-uniformity device noise such as blind pixel and stripe noises). Therefore, targets usually exhibit a low signal-to-clutter ratio (SCR), and are easily immersed in various clutters and noise. 
	\item \textbf{Various compound satellite motion:} The presence of translation, pitch, yaw, roll, jitter, and satellite scheduling significantly degrades imaging quality, leading to diminished target intensity and blurred target outlines. In addition, the coupled target and background motion pose challenges for motion information extraction and utilization. {Since existing methods mainly focus on scenarios with a relatively static or slightly modified field of view, they encounter difficulties in handling the sophisticated motion characteristics of satellite video.}
	\item \textbf{Lack of open-source datasets:} The lack of open-source MIRST datasets in satellite videos greatly hinders the advancement of MIRST detection algorithms. Therefore, a large-scale, high-quality dataset is absolutely necessary.
\end{itemize}

{Dataset is the foundation underlying research development. Therefore, we build the first large-scale MIRST dataset in satellite videos (namely the IRSatVideo-LEO dataset), including 200 sequences, 91366 frames with mask annotations.} Due to the limited availability of infrared satellite videos, IRSatVideo-LEO is a semi-simulated dataset with a real satellite image and synthesized satellite motion, target appearance, trajectory and intensity, which can provide a standard and efficient toolbox for satellite video generation, and offers a reliable evaluation platform to facilitate the algorithms development. Note that, in this paper, we mainly focus on LEO and GEO satellites with smoothly moving satellite platforms, while the  GEO satellite videos can also be simulated by our toolbox.

Based on the IRSatVideo-LEO dataset, we develop a recurrent feature refinement (RFR) framework as the baseline method for MIRST in satellite videos. RFR can be equipped with existing deep learning-based single-frame infrared small target (SIRST) detection methods for joint satellite motion compensation and MIRST detection in a data-driven manner. Specifically, RFR employs recurrent framework \cite{deng2021rfrn,haris2019recurrent} to fully exploit the long-term temporal dependency from the entire input video. Through RFR, features are iteratively propagated and aggregated, which retains context information from all history memory to refine the current state. 
For feature propagation with moving platform, we employ deformable convolutions \cite{DCN,DCNv2,TDAN} to perform implicit motion compensation. Note that, considering the scale differences between background motion and target motion, we use a pyramid structure to decompose the motion, and perform target and background motion compensation in a coarse-to-fine manner. 
For feature aggregation, we propose a temporal-spatial-frequency modulation (TSFM) module to aggregate the beneficial temporal information from aligned features, and adaptively enhance the spatial and frequency information especially for small targets. 
In summary, the contributions of this paper can be summarized as follows:

\begin{itemize}
	\item {This paper, for the first time, discusses the multi-frame infrared small target detection in satellite videos (MIRST-SatVideo), and builds the first large-scale dataset (namely IRSatVideo-LEO) to lay the foundation of research development.}
	\item {Based on IRSatVideo-LEO, we develop a recurrent feature refinement (RFR) framework as a baseline method for MIRST detection in satellite videos, which can fully exploit the long-term temporal dependency to achieve satellite motion compensation and MIRST detection in an end-to-end manner.}
	\item {Extensive experiments have been conducted to demonstrate the effectiveness and superiority of our method. In addition, RFR can be easily equipped with existing SIRST detection algorithms to achieve consistent performance improvements.}
\end{itemize}

This paper is organized as follows: Section \ref{sec-related} briefly reviews the related work. Section \ref{Dataset} introduces our self-developed IRSatVideo-LEO dataset in detail. Section \ref{Method} introduces the architecture of our method. The experimental results are presented in Section \ref{experiments}. Section \ref{sec-conclusion} gives the conclusion.

\section{Related Work}\label{sec-related}

In this section, we briefly review the major works of algorithms and datasets for IRST detection.

\subsection{Algorithms for IRST Detection}

IRST Detection is a long-standing, fundamental yet challenging task in infrared search and tracking systems. Due to the rapid reaction of high maneuvering target, numerous single-frame IRST detection methods have been proposed in the past decades, including early traditional paradigms (\textit{e.g.,} filtering-based methods \cite{tophat,Max-Median}, local contrast-based methods \cite{LCM,RLCM,MSPCM,WSLCM,TLLCM}, low rank-based methods \cite{RIPT,NRAM,ASTTV-NTLA,PSTNN,MSLSTIPT}) and recent deep learning paradigms \cite{MDvsFA,ACM,ALCNet,DNANet,ISNet,RKformer,FC3-Net,ISTDet,RISTDnet,AGPCNet,RPCANet}.

{Based on the aforementioned single-frame methods, temporal clues within sequence images can be exploited \cite{3Dfilter0,ASTTV-NTLA,DTUM,SRSTT} to address blur, distortion and low-contrast targets within a single frame, and can offer reasonable history tracking and future forecasting. Specifically, a variety of spatio-temporal filters (\textit{e.g.,} 3-D matched filter \cite{3Dfilter0,3Dfilter1,3Dfilter2}, correlation filter \cite{Corrfilter0,Corrfilter1}, particle filter \cite{particleFilter0,particleFilter1}, wavelet filter \cite{waveletFilter0,waveletFilter1,waveletFilter2}) are proposed to suppress low-frequency background and extract the region of interest. Recently, incorporating spatio-temporal information into local contrast-based methods \cite{Li2016,zhao2023adaptive,STVDM} and low rank-based methods \cite{MSLSTIPT,ASTTV-NTLA,IMNN-LWEC,SRSTT,STTM-SFR,CTSTC} have raised more and more attention. Although these series of methods perform robust to low-contrast targets, their handcrafted features, fixed hyper-parameters and high computational complexity cannot generalize well to real scenes with complex background clutter and real-time application requirements. To address this problem, powerful CNN-based methods \cite{DTUM,3DSTPM,STDMANet,DTNet} have emerged to learn trainable features in a data-driven manner, which can achieve state-of-the-art performance with high efficiency. Specifically, Li \textit{et. al.} designed a direction-coded convolution block to distinguish the motion between targets and clutters. Zhang \textit{et. al.} incorporated a spatial-temporal tensor (STT) optimization model into CNN-based method to achieve IRSTD in a model \& data-driven manner. Yan \textit{et. al.} proposed a spatio-temporal differential multi-scale attention network (STDMANet) to extract temporal multi-scale features for IRSTD. However, these methods only focus on target motion under a relatively static field of view (some are with sudden movement by turnable collection), which cannot generalize well to the complex background motion of satellite videos.}

{Recently, motion estimation and compensation have been widely used in many video processing tasks, such as video super-resolution \cite{qing2023video,wang2020deep,ying2020deformable}, video inpainting \cite{xu2019deep,wu2021dapc}, video object detection \cite{zhu2017flow,zhou2022transvod,HiEUM}. Compared with two-step optical flow-based approaches \cite{wang2020deep,xu2019deep,zhu2017flow} (\textit{i.e.,} motion estimation by optical flow approaches and frame alignment by warping) that always leads to ambiguity and duplication, deformable convolution-based methods \cite{ying2020deformable,wu2021dapc} achieve motion compensation in a unified step by dynamically adjusted sampling grid in a learnable manner. Therefore, we employ deformable convolution for satellite motion compensation in our paper.}
	
\begin{figure*}[t]
	\centering\includegraphics[width=18cm]{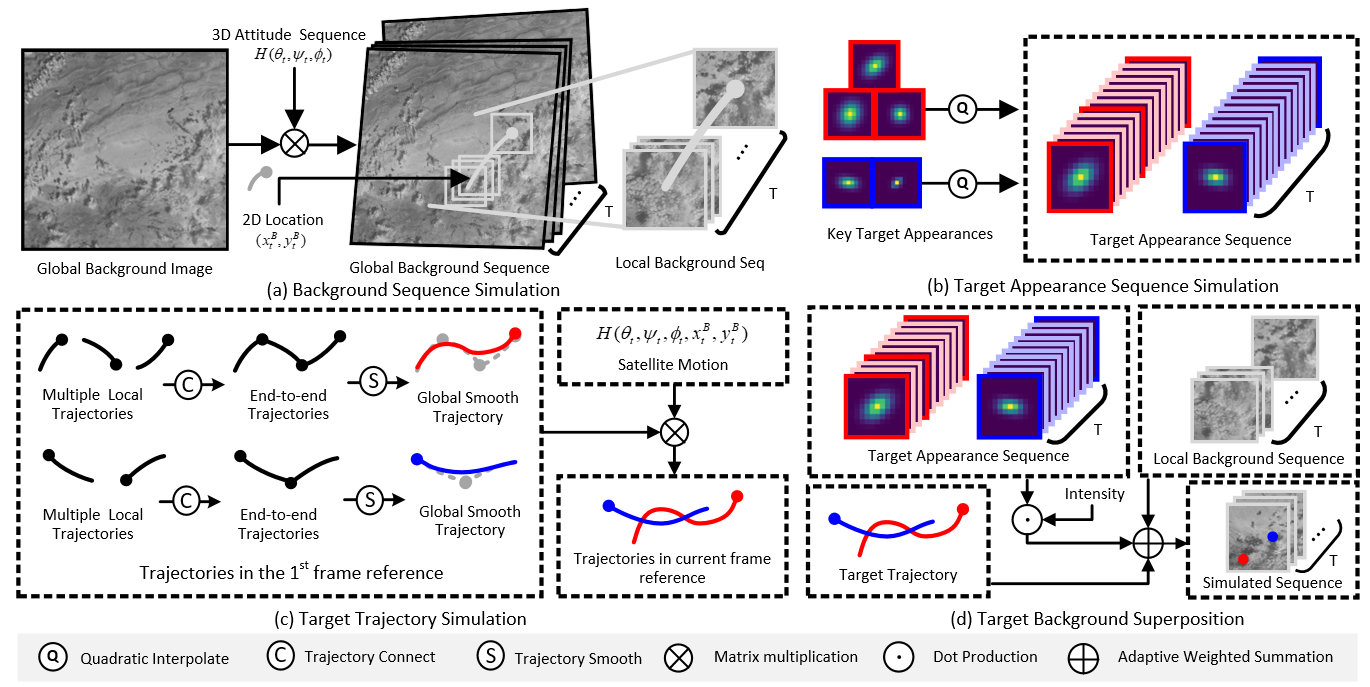}
	\caption{Implementation details of the IRSatVideo-LEO dataset that consists of four steps: (a) Background sequence simulation. 3D attitude sequence is first used to generate a global background sequence from a SWIR satellite image by homography transformation, and 2D location is then used to crop the local background sequence. (b) Target appearance sequence simulation. Several key target appearances (\textit{e.g.,} red and blue Gaussian kernel images) are used to generate target appearance sequence by quadratic interpolation\textsuperscript{\ref{web}} (\textit{e.g.,} light red and light blue Gaussian kernel images are interpolated results). (c) Target trajectory simulation. We first generate the trajectory in the $1^{st}$ frame reference by connecting and smoothing multiple separate trajectories, and then employ satellite motion to perform reference transformation to generate trajectory in the current frame reference. (d) Target background superposition. Target appearance sequence and intensity are multiplied, which is then adaptive weighted summed by background sequence using target trajectory to generate the simulated sequence.}\label{fig-simulation}
	\vspace{-.3cm}
\end{figure*}

\subsection{Datasets for IRST Detection}

Due to special applications \cite{NFTDGSTV,Hou2024,ASTTV-NTLA}, IRST datasets are always scarce. In recent decades, plenty of SIRST detection datasets \cite{IPI,MDvsFA,ACM,DNANet,ISNet,NUDT-SIRST-Sea} and multi-frame IRST (MIRST) detection datasets \cite{Hui1,Hui2,SIATD,DTUM,RDIAN} have been proposed to facilitate the algorithm development. 

As the pioneering open-source SIRST detection datasets, Wang \textit{et. al.} \cite{MDvsFA} synthesized 10100 images with real scenes and simulated targets. Then, Dai \textit{et. al.} \cite{ACM} collected and annotated 427 images to release the first real dataset (namely SIRST), which was then extended to SIRST-v2 with 514 images \cite{OSCAR}. To further enlarge the available dataset, the simulated NUDT-SIRST dataset and real IRSTD-1K dataset are proposed for over a thousand images. Besides the aforementioned land-based and aerial-based SIRST datasets, Wu \textit{et. al.} \cite{NUDT-SIRST-Sea} proposed the first space-based near-infrared tiny ship dataset mounted by a low Earth-orbiting satellite. 

Compared with SIRST detection datasets that only provide appearance information, consecutive frames of MIRST detection datasets allow for a more comprehensive understanding of target behavior by additional motion patterns. In addition, the inherent appearance \& radiation variations, motion blur, dynamic background and various annotation forms (\textit{e.g.,} point, box and mask) of MIRST detection datasets significantly increase the difficulty of this task. Hui \textit{et. al.} \cite{Hui1} and Fu \textit{et. al.} \cite{Hui2} released two real MIRST detection datasets through the 2$^{nd}$ and 3$^{rd}$ Sky Cup competitions, which contain 22 and 87 sequences with point and box annotations, respectively. Sun \textit{et. al.} \cite{SIATD} semi-simulated 350 image sequences (namely SIATD) with background sequences captured by UAVs and simulated target appearance and motion. SIATD only provides point annotations. Sun \textit{et. al.} \cite{RDIAN} built a large-scale infrared small dim target dataset (IRDST). IRDST contains a real subset of 85 sequences and a simulated subset of 316 sequences with pixel-level annotations. Li \textit{et. al.} \cite{DTUM} synthesized a multi-frame infrared small and dim target dataset (NUDT-MIRSDT) to explore MIRST detection under extremely low signal-to-noise ratio (\textit{i.e.,} SNR$<$3).

Despite the valuable contributions of open-source datasets in the field of MIRST detection, these datasets are all captured by aerial-based and land-based imaging systems, and have large differences against space-based datasets (see section \ref{sec-intro} for details). Due to the scarcity and unavailability of infrared satellite videos, we decide to develop the first large-scale MIRST dataset in satellite videos, namely the IRSatVideo-LEO dataset, to advance the technique development of space-based MIRST detection.

\section{The IRSatVideo-LEO Dataset}\label{Dataset}
Large quantity, rich diversity, and high-quality labeled data are the cornerstone of deep learning-based algorithms, especially with the emergence of large-scale models. However, the inherent data scarcity of infrared satellite videos and high confidentiality of targets (\textit{e.g.,} aircraft, missile, and launch vehicle) greatly hinder the development of MIRST detection on satellite videos. Inspired by other data-scarcity fields \cite{SIATD,Data_scarcity1,Data_scarcity2,DSFnet,SatelliteVOD,RGBT-Tiny}, we develop a large-scale dataset (\textit{i.e.,} IRSatVideo-LEO) using semi-simulated implementation with a real satellite image and synthesized satellite motion, target appearance, trajectory and intensity. Please refer to section \ref{Dataset_Implementation_Details} for more details. 

\subsection{Background Image Collection}
\begin{figure}[t]
	\centering\includegraphics[width=9cm]{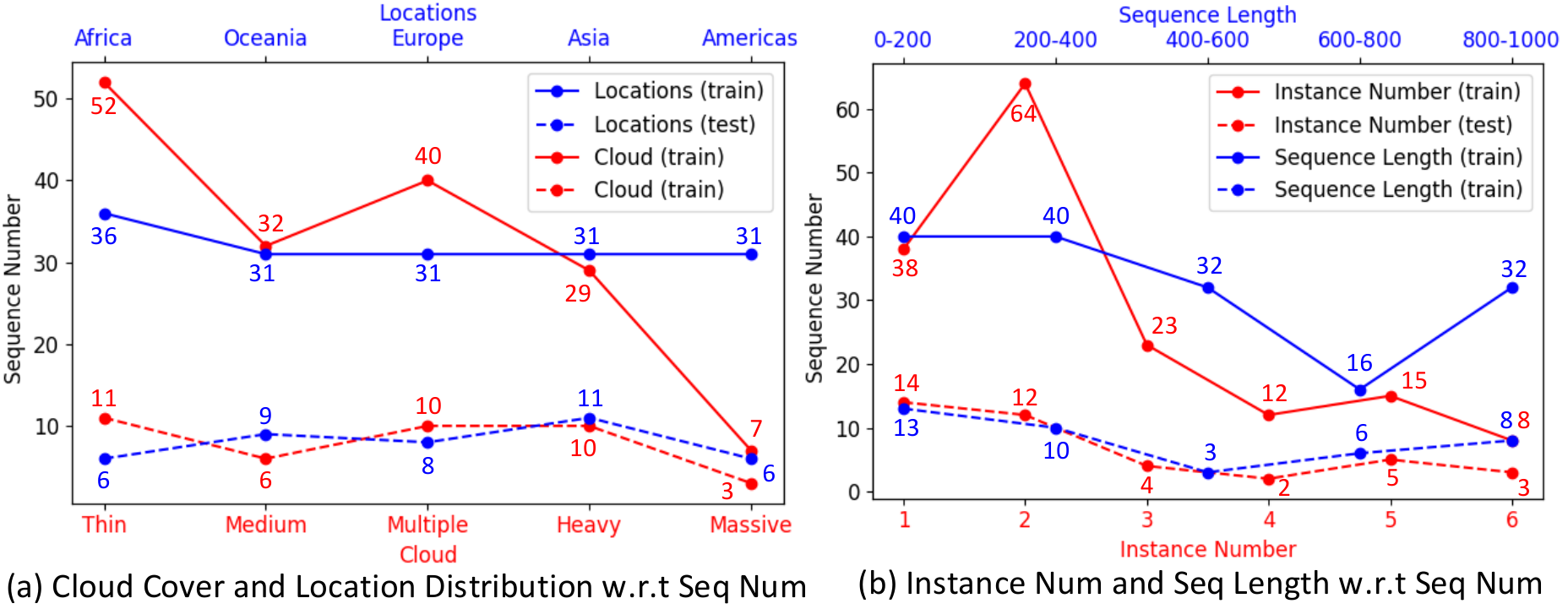}
	\vspace{-.4cm}
	\caption{{Illustrations of sequence attributes. (a) shows cloud cover \& location distribution with respect to (w.r.t) sequence number in training and test dataset. (b) shows instance number \& sequence length with w.r.t sequence number in training and test dataset. Numbers represent the corresponding sequence number.}}\label{fig:seq_attribute}
	\vspace{-.2cm}
\end{figure}

We collect 200 background images captured by the 7$^{th}$ band (\textit{i.e.,} SWIR 2.1-2.3 $\upmu m$) of Landsat 8 and 9. To ensure the generalization of the dataset, we randomly sample locations across each continent and ocean on Earth, and the cloud cover ratios range from 0 to 61.25\%. {Figure \ref{fig:seq_attribute} shows the cloud cover ratios and the location distributions of the IRSatVideo-LEO dataset.} 

\subsection{Implementation Details}\label{Dataset_Implementation_Details}

As shown in Fig.~\ref{fig-simulation}, to render an image sequence from a single satellite image, four steps are required: background sequence simulation, target appearance sequence simulation, target trajectory simulation and target background superposition. Then, we introduce these four steps in detail. {Note that, the parameters used for data generation have been listed in Table~\ref{tab-hyparameters}. Among them, different parameters result in different satellite motion, target appearance, trajectory, intensity, and thus enrich the data varieties. All parameters can adapt to accommodate more scenarios.}

\subsubsection{Background Sequence Simulation}

Moving background results from satellite motion, which is composed of six degrees of freedom: 3D attitudes (\textit{i.e.,} pitch $\alpha$, yaw $\beta$, and roll $\gamma$ angles) and 3D locations (\textit{i.e.,} translation along x, y and z axes). 
As shown in Fig.~\ref{fig-simulation} (a), we first employ the homography of 3D attitude sequence to warp the global background image to generate the global background sequence. The process can be formulated as:
{\begin{align}
	I_t^{GB} = I^{GB}\otimes \textbf{H}(\alpha_{t},\beta_{t},\gamma_{t}),
\end{align}}
{where $I^{GB}$$\in$$\mathbb{R}^{H_0\times W_0}$ is the global background image, and $I_t^{GB}$ is the global background sequence. $t$$\in$$[1,T]$ and $T$ is the sequence length. $\alpha_{t},\beta_{t}, \gamma_{t}$ represent the pitch, yaw and roll angles at time $t$. $\textbf{H}(\cdot)$ represents the homography transformation, and $\otimes$ represents matrix multiplication. For satellite without scheduling (\textit{i.e.,} constant attitude), we randomly set fixed 3D attitudes $(\alpha_{t},\beta_{t},\gamma_{t})$=$($a$,$b$,$c$)$, a$\in$$[-10,10]$, b$\in$$[-10,10]$, c$\in$$[-10,10]$. For satellite with scheduling (\textit{i.e.,} attitude changes slowly), we set the initial $(\alpha_{1},\beta_{1},\gamma_{1})$ and final $(\alpha_{T},\beta_{T},\gamma_{T})$ 3D attitudes, and employ linear interpolation to generate intermediate attitudes. Note that, scheduling range $(\Delta\alpha,\Delta\beta,\Delta\gamma)$=$(\alpha_{T},\beta_{T},\gamma_{T})$-$(\alpha_{1},\beta_{1},\gamma_{1})$ are randomly sampled from $[-5,5]$, and the probability of these two modes (\textit{i.e.,} constant attitudes, and slowly changed attitudes) are both set to $0.5$. }

\begin{figure}[t]
	\centering\includegraphics[width=9cm]{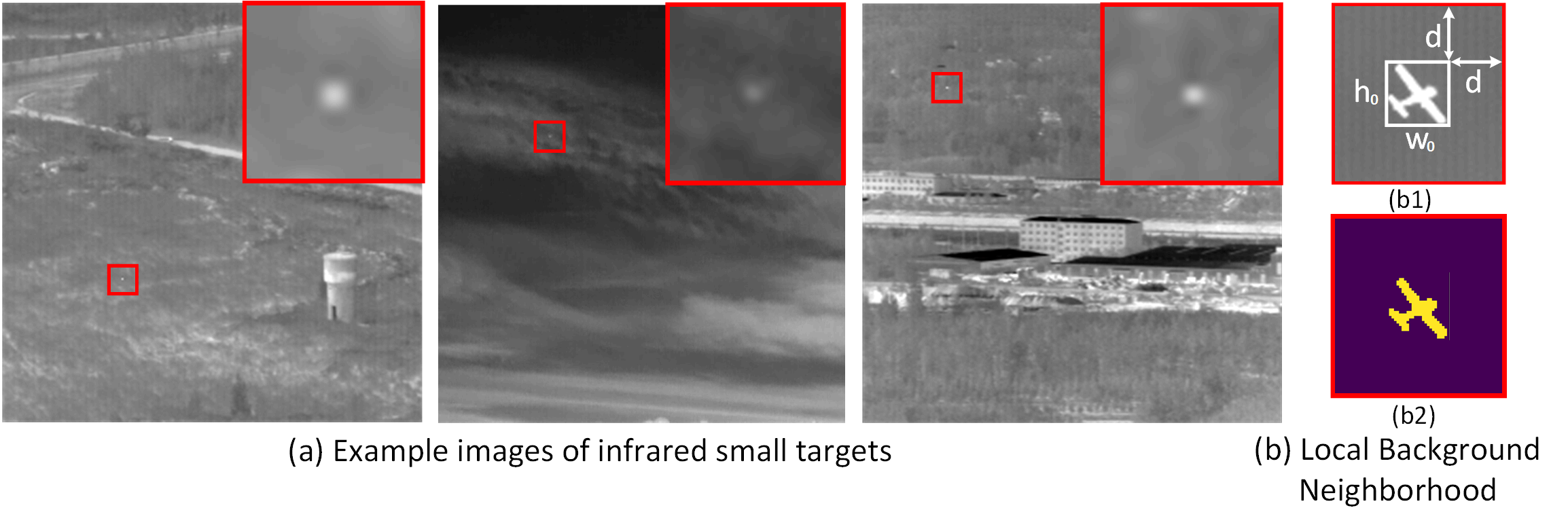}
	\vspace{-.65cm}
	\caption{Illustrations of infrared small target and local background neighborhood. (a) Example images of infrared small targets. (b1) Local background neighborhood is extended from the BBox of the target ($h_0$ in height and $w_0$ width) by $d$ in both height and width. (b2) illustrates the target region (\textit{i.e.,} yellow area) and local background region (\textit{i.e.,} purple area).}\label{fig-Guassian}
\end{figure}

For 3D satellite locations, translation along the z-axis is ignored due to the space-based imaging system. To generate the 2D satellite location sequence $(x_t^B,y_t^B)$, we randomly set the initial $(x_1^B,y_1^B)$ and final $(x_T^B,y_T^B)$ 2D locations, and employ quadratic interpolation\footnote{Given the 2D initial $(x_1, y_1)$ and final $(x_T, y_T)$ points, we randomly extract a segment of a low-order curve with order $p\in (0,3)$, length $x_T-x_1$ along x-axis, and $y_T-y_1$ along y-axis.\label{web}} for intermediate locations. {Note that, the velocities along x-axis and y-axis $v_x^B$=$\frac{x_T^B-x_1^B}{T}$, $v_y^B$=$\frac{y_T^B-y_1^B}{T}$ are randomly sampled from $[1/20,2]$. Then local background sequence is cropped according to the 2D satellite location sequence, and the pre-defined field of view. The process can be formulated as:}
{\begin{align}
I_t^{LB} = Crop(I_t^{GB}, x_t^B, y_t^B, H_0, W_0),
\end{align}}
{where $I_t^{LB}$$\in$$\mathbb{R}^{H_0\times W_0}$ represents the local background sequence. $Crop$ represents the crop the global background sequence $I_t^{GB}$ by 2D satellite location $x_t^B, y_t^B$ with a filed of view of $H_0, W_0$, and $H_0$ and $W_0$ in IRSatVideo-LEO are both set to 1024 \cite{NUDT-SIRST-Sea}.}  Note that, our paper focuses on medium and low orbit satellites (\textit{i.e.,} smoothly moving background), and you can set $(\alpha_t,\beta_t,\gamma_t,x_t^B,y_t^B)$ to zero or small random values for geostationary orbit satellite (\textit{i.e.,} static or slightly jittering background). Figure~\ref{fig-BM_TS} (a1), (b1) shows two examples of local background sequence in the $1^{st}$ frame reference. It can be observed that satellite motion exhibits random, slow and smooth, which is in accordance with the real-world scenario.

\begin{figure}[t]
	\centering\includegraphics[width=9cm]{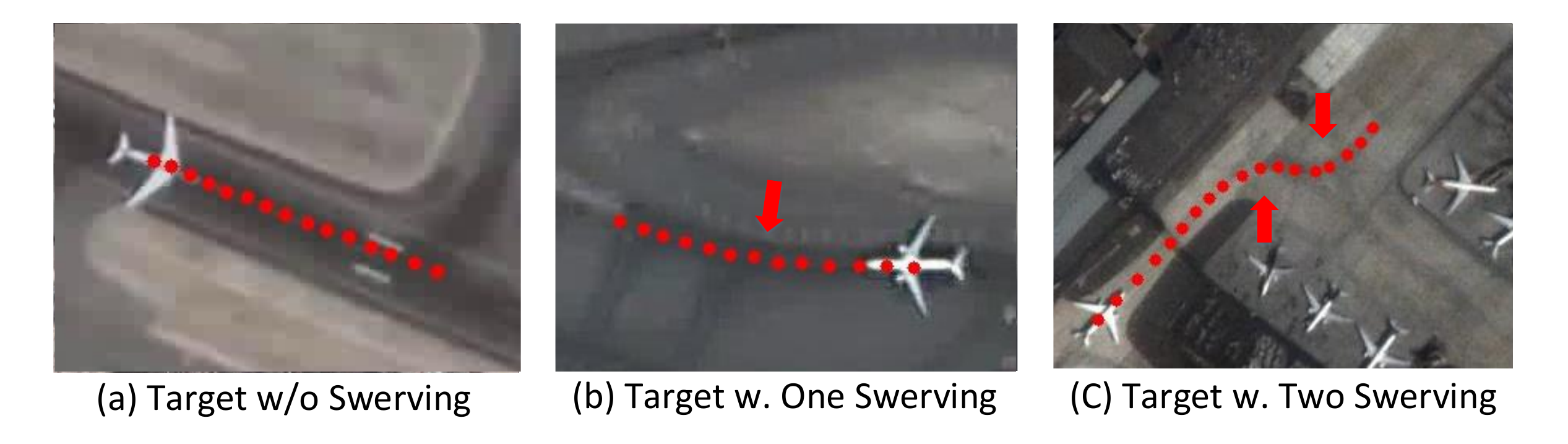}
	\vspace{-.6cm}
	\caption{Example target trajectories in satellite videos. (a), (b), (c) shows the trajectory of target without (w/o) and with (w.) one and two swerving. Note that, discrete points of target trajectory with a sampling rate of 20 are shown for better visualization. The density of discrete points represents the velocity of the target, and dense points represent high velocity. Red arrows are specific to the swerving position.}\label{fig-Target_Trajectory} %
	\vspace{-.4cm}
\end{figure}

\begin{table*}[t]
	\begin{center}
		\centering
		\renewcommand\arraystretch{1.45}
		\caption{{Details of parameters for data generation. ``BSS", ``TAS", ``TTS" and ``TBS" represent background sequence simulation, target appearance simulation, target trajectory simulation and target background superposition.}}\label{tab-hyparameters}
		\footnotesize
		{\begin{tabular}{ccccc}
				\toprule[1pt]
				&Symbol&Description&Distribution&Value\\
				\midrule
				\multirow{12}*{\rotatebox{90}{BSS}}
				&$t$&time&Linear&$t$=$\{1,2,...,T$-$1,T\}$, $T$$\in$$[200,1200]$\\
				&\multirow{2}*{$\alpha_t$}&\multirow{2}*{Pitch angle at time t}&Constant&$\alpha_t$=$a$, $a$$\in$$[-10,10]$\\
				&&&Linear&$\Delta\alpha$=$\alpha_{T}-\alpha_{1}$,  $\Delta\alpha\in$[-5,5]\\
				&\multirow{2}*{$\beta_t$}&\multirow{2}*{Yaw angle at time t}&Constant&$\beta_t$=$a$, $a\in$$[-10,10]$\\
				&&&Linear&$\Delta\alpha$=$\beta_{T}-\beta_{1}$,  $\Delta\beta\in$[-5,5]\\
				&\multirow{2}*{$\gamma_t$}&\multirow{2}*{Roll angle at time t}&Constant&$\gamma_t$=$a$, $a\in$$[-10,10]$\\
				&&&Linear&$\Delta\gamma$=$\gamma_{T}-\gamma_{1}$,  $\Delta\gamma\in$[-5,5]\\
				&$x_t^B$&Translation along x-axis at time t&Quadratic&$v_x^B=(x_T^B-x_1^B)/T$, $v_x^B\in$$[1/20,2]$\\
				&$y_t^B$&Translation along y-axis at time t&Quadratic&$v_y^B=(y_T^B-y_1^B)/T$, $v_y^B\in$$[1/20,2]$\\
				&$z_t^B$&Translation along z-axis at time t&Constant&$0$\\
				&$H_0$&Height of field of view&Constant&$1024$\\
				&$W_0$&Width of field of view&Constant&$1024$\\
				\midrule
				\multirow{5}*{\rotatebox{90}{TAS}}
				&$h_t$&Major axis length of Gaussian target at time t&Multiple Quadratic&$h$$\in$$[1,9]$\\
				&$w_t$&Minor axis length of Gaussian target at time t&Multiple Quadratic&$w$$\in$$[1,9]$\\
				&$\sigma_t$&Sigma value of Gaussian target at time t&Multiple Quadratic&$\sigma$$\in$$[0.1,1]$\\
				&$N$&Number of Gaussian targets&Constant&$N$$\in$$\{1,2,3,4,5,6\}$\\
				&$K$&Number of intermediate states of a Gaussian target&Constant&$K$$\in$$[2,3,4,5]$\\
				\midrule
				\multirow{6}*{\rotatebox{90}{TTS}}
				&$x_{nt}^T$&Location x in the $1^{st}$ frame reference at time t&Multiple Smoothed Low-order Curves&-\\
				&$y_{nt}^T$&Location y in the $1^{st}$ frame reference at time t&Multiple Smoothed Low-order Curves&-\\
				&$\hat{x}_{nt}^T$&Location x in the current frame reference at time t&Multiple Smoothed Low-order Curves&-\\
				&$\hat{y}_{nt}^T$&Location y in the current frame reference at time t&Multiple Smoothed Low-order Curves&-\\
				&p&Order number of curve&Constant&$p$$\in$$[0,3]$\\
				&S&Target swerving times&Constant&$S$$\in$$\{0,1,2\}$\\
				\midrule
				\multirow{3}*{\rotatebox{90}{TBS}}
				&$scr$&SCR value of a target at time 1&Constant&$scr$$\in$$[1,20]$\\
				&$k_{gb}$&kernel size of Gaussian blur&Constant&$k_{gb}$$\in$$\{3,5,7\}$\\
				&$\sigma_{gb}$&Sigma of Gaussian blur&Constant&$\sigma_{gb}$$\in$$[0.2,0.6]$\\
				\bottomrule[1pt]
		\end{tabular}}
	\end{center}
\end{table*}

\begin{table*}
	\footnotesize
	\centering
	\begin{threeparttable}
		\vspace{-.4cm}
		\caption{{Statical comparisons among existing SIRST and MIRST detection datasets and our IRSatVideo-LEO dataset. ``Seq." and ``Frame" represent the number of sequences and frames. ``T-Num.", ``T-Size" and ``T-SCR" represent the target number, average target size and average target SCR, respectively. Note that, ``T-Size" is the pixel number with mask annotation, while is the box area with BBox annotation. ``T-SCR" with BBox or point annotation ($^*$) is calculated by the average value of SCR and SNR (\textit{i.e.,} replacing $\mu_t$ in Eq.~\ref{eq-SCR} by the maximum value in target region).}}\label{tab-Statics}
		\setlength{\tabcolsep}{1.8mm}
		{
		\begin{tabular}{|clcccccccccc|}
			\hline
			&Benchmark&Image Type&Scene Type&Label Type&Wave Band&Resolution&Seq.&Frame&T-Num.&T-Size&T-SCR\\\hline
			\multirow{5}*{\rotatebox{90}{SIRST}}&NUST-SIRST \cite{MDvsFA}&Synthetic&Land&Mask&Thermal&129$\times$129$^1$&-&10100&10337&40&5.40\\
			&SIRST-v2 \cite{OSCAR}&Real&Aerial&Mask&Thermal&278$\times$366$^2$&-&514&648&37&12.07\\
			&NUDT-SIRST \cite{DNANet}&Synthetic&Aerial/Land&Mask&Thermal&256$\times$256&-&1327&1863&34&5.60\\
			&IRSTD-1K \cite{ISNet} &Real&Land&Mask&Thermal&512$\times$512&-&1001&1492&53&8.92\\
			&NUDT-SIRST-Sea \cite{NUDT-SIRST-Sea}&Real&Space&Mask&Near&998$\times$998$^3$&-&5808&16929&36&15.32\\\hline
			\multirow{9}*{\rotatebox{90}{MIRST}}
			&Hui \cite{Hui1} &Real&Land&Point&Thermal&256$\times$256&22&16177&16944&$-$&5.33$^*$\\
			&Fu \cite{Hui2}&Real&Aerial&BBox&Thermal&640$\times$480&87&21750&89174&185&4.66$^*$\\
			&Anti-UAV v2 \cite{Anti-UAV}&Real&Land&BBox&Thermal&640$\times$512&140&152561&152561&3219&1.32$^*$\\
			&IRDST-Real \cite{RDIAN}&Real&Land&Mask&Thermal&992$\times$742&85&40656&41801&10&6.89\\
			&IRDST-Simulation \cite{RDIAN}&Synthetic&Land&Mask&Thermal&720$\times$480&316&106254&102077&6&4.68\\
			&SIATD \cite{SIATD} &Synthetic&Aerial&Point&Thermal&640$\times$512&350&150185&247080&$<$7$\times$7&6.95$^*$\\
			&NUDT-MIRSDT \cite{DTUM} &Synthetic&Aerial&Mask&Thermal&217$\times$302$^4$&120&12000&11464&34&1.87\\
			&{SIRSTD \cite{ST-Trans}} &{Real}&{Land}&{BBox}&{Thermal}&{640$\times$512}&{48}&{50388}&{48565}&{60}&{10.40}\\
			&IRSatVideo-LEO &Synthetic&Space&Mask&Short Wave&1024$\times$1024&200&91021&220126&4&6.69\\\hline
		\end{tabular}}
		\begin{tablenotes}[flushleft] 
			\scriptsize 
			\item[{$1234$}] {Varied image resolution in NUST-SIRST dataset of $h$$\in$$[96,327]$, $w$$\in$$[101,442]$, SIRST-v2 dataset of $h$$\in$$[96,1024]$, $w$$\in$$[135,1280]$, NUDT-SIRST-Sea dataset of $h$$\in$$[740,1024]$, $w$$\in$$[740,1024]$, NUDT-MIRSDT dataset of $h$$\in$$[154,324]$, $w$$\in$$[209,407]$. The average image resolution is shown for simplicity.}
		\end{tablenotes}
	\end{threeparttable}
\vspace{-.2cm}
\end{table*}

\subsubsection{Target Appearance Simulation}

Due to the remote sensing imaging systems and optical diffraction effect, infrared small targets always appear as small points or diffraction spots.  {As shown in Fig.~\ref{fig-Guassian} (a), targets can be approximated by Gaussian kernels $G(h,w,\sigma)$ \cite{Gaussian_target1,Gaussian_target2,SIATD}, where $h$, $w$ are major \& minor axis, and $\sigma$ represents sigma value. We randomly simulate $N$$\in$$[1,6]$ Gaussian targets for a image sequence, and $h$$\in$$[1,9]$, $w$$\in$$[1,9]$, $\sigma$$\in$$[0.1,1]$. To generate the target appearance sequence of the $n^{th}$ target $G_{nt}(h_t,w_t,\sigma_t)$, as shown in Fig.~\ref{fig-simulation} (b), we randomly set $K$$\in$$[2,5]$ Gaussian kernels $G_{nt_1}$, $G_{nt_2}$,...,$G_{nt_K}$ ($t_1$=$1$, $t_K$=$T$, $\{t_2$...$t_{K-1}\}$$\in$$(1,T)$) as the key target appearances in a sequence, and employ quadratic interpolation\textsuperscript{\ref {web}} to generate intermediate appearances.} 

\subsubsection{Target Trajectory Simulation}

\begin{figure}[t]
	\vspace{.1cm}
	\centering\includegraphics[width=9cm]{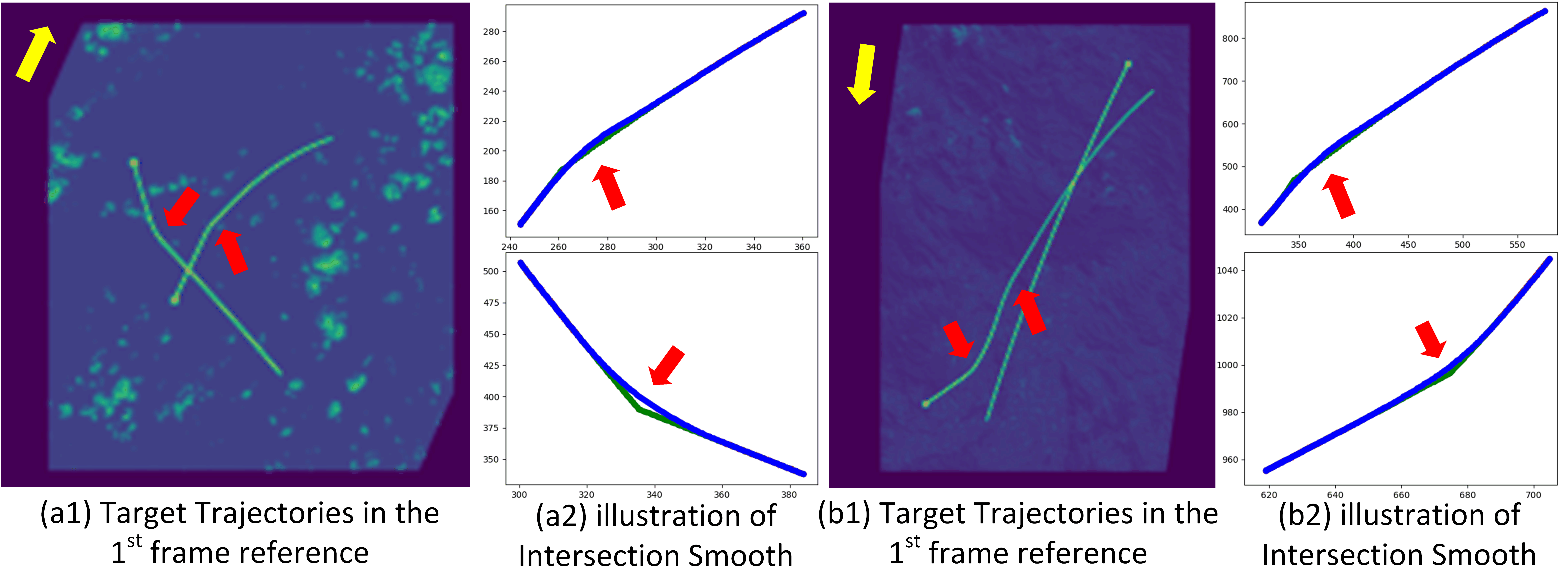}
	\caption{Illustrations of moving background and swerving target. For moving background, (a1) and (b1) show image sets of the local background sequence in the $1^{st}$ frame reference, and the yellow arrows represent the moving directions of background. For swerving target, (a1) and (b1) show the target trajectories in the $1^{st}$ frame reference, and the end of trajectories are labeled by circular arrows. (a2) and (b2) show the original end-to-end adjacent trajectories (\textit{i.e.,} green lines) and the smooth adjacent trajectories (\textit{i.e.,} blue lines). Red arrows are specific to the swerving position.}\label{fig-BM_TS}
	\vspace{-.4cm}
\end{figure}

Figure~\ref{fig-Target_Trajectory} shows some examples of target trajectories in satellite video sequences \cite{SatVideoDT}. It can be observed that the real target trajectory is continuous and relatively smooth over a short time interval, which can be approximated using a low-order curve with order $p$$\in$$(0,3)$ \cite{SIATD}. In addition, the target may swerve to change the moving direction and velocity. To simulate the $n^{th}$ target trajectory with $S$$\in$$[0,2]$ swerving times, as shown in Fig.~\ref{fig-simulation} (c), we first synthetic $S$ local target trajectories by low-order curves, and sequentially connect them end-to-end. Then, we employ 1D interpolation to smooth each intersection of two adjacent trajectories for a global smooth trajectory $(x_{nt}^T,y_{nt}^T)$. Figure~\ref{fig-BM_TS} shows the trajectories of two swerved targets in the $1^{st}$ frame reference. It can be observed that target trajectories are locally continuous and globally smooth, which are in accordance with the real-world target motion. It is worth noticing in Fig.~\ref{fig-BM_TS} (b1) that, the moving direction of one target is opposite to that of the background, which indicates that the target disappears in some frames of the simulated sequence.

With satellite motion, we employ homography transformation to generate the target locations in the image plane. That is, the target trajectories in the $1^{st}$ frame reference are wrapped by the homography matrix of satellite motion to generate the one in current frame reference. The process can be formulated as:
\begin{align}
	(\hat{x}_{nt}^T,\hat{y}_{nt}^T) = (x_{nt}^T,y_{nt}^T)\otimes \textbf{H}(\alpha_{t},\beta_{t},\gamma_{t},x_t^B,y_t^B),
\end{align}
where $(\hat{x}_{nt}^T,\hat{y}_{nt}^T)$ is the target trajectory in current frame reference, and $(\alpha_{t},\beta_{t},\gamma_{t},x_t^B,y_t^B)$ are the parameters of satellite motion.

\subsubsection{Target Background Superposition}
{As shown in Fig.~\ref{fig-simulation} (d), we perform dot production between the appearance sequences $G_{nt}$ and intensity sequences $E_{nt}$ of the $n^{th}$ target to generate the target template sequences $I_{nt}$. The formulation can be formulated as:}
{\begin{align}
	I_{nt}&=G_{nt}\odot E_{nt},\\
	E_{nt}&= [scr\times \sigma(T_1^{LB}) + \mu(T_1^{LB})]  \times (1+a_{nt}),
\end{align}}
{where $\mu(T_1^{LB})$ and $\sigma(T_1^{LB})$ represents the mean and standard deviation value of target local background at time 1, and $scr$ is randomly sampled from $[1,20]$. $a_{nt}$ represents the target accelerate sequence, and $\odot$ represents element-wise multiplication. Finally, we perform adaptive weighted summation \cite{Background_target_embedding,SIATD} between the target template sequence and local background sequence, and impose a Gaussian blur function \cite{DNANet} for smoothness. The process can be formulated as:}
{\begin{align}
	\hat{I}_{nt}^{sim}&=GaussianBlur(I_{nt}^{sim}, k_{gb}, \sigma_{gb}),\\
	I_{nt}^{sim}&=Norm(I_{nt}^{T})\odot I_{nt}^{T}+(1-Norm(I_{nt}^{T}))\odot I_{nt}^{LB},
\end{align}}
{where $\hat{I}_{nt}^{sim}$ and $I_{nt}^{LB}$ represent the image patch in location $(\hat{x}_{nt}^T,\hat{y}_{nt}^T)$ of simulated sequence and local background sequence, respectively. Since location $(\hat{x}_{nt}^T,\hat{y}_{nt}^T)$ is generally fractional, we follow \cite{Background_target_embedding,SIATD} to use bilinear interpolation to generate exact values. $Norm(I)$$=$$I/max(I)$ represents image normalization, and $max(I)$ represents the maximum value of the image. $GaussianBlur$ represents the Gaussian blur function with kernel size $k_{gb}$$\in$$\{3,5,7\}$ and sigma $\sigma_{gb}$ randomly sampled from $[0.2,0.6]$.}

\subsection{Training-test Sets and Annotations}

\begin{figure}[t]
	\centering\includegraphics[width=9cm]{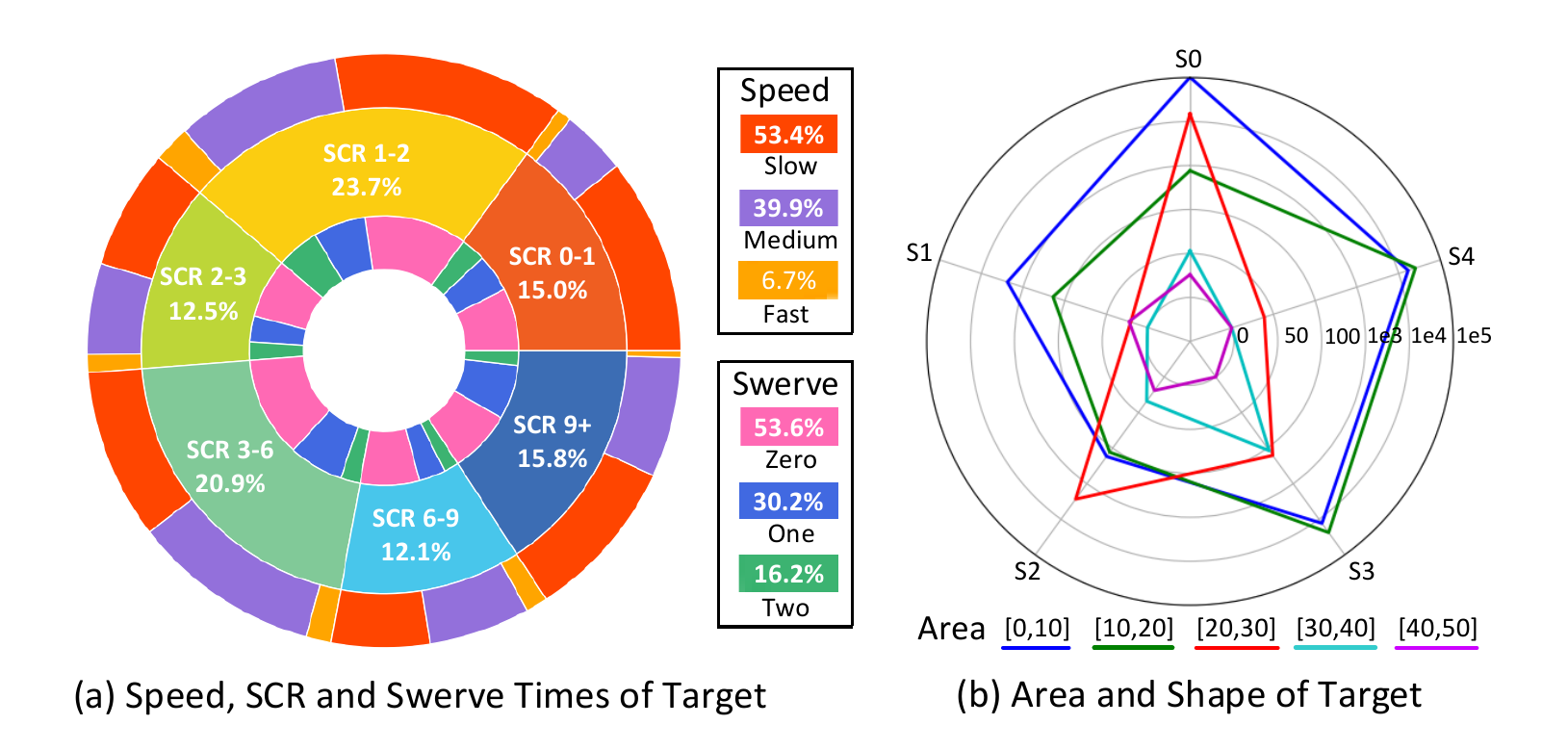}
	\vspace{-.4cm}
	\caption{{Illustrations of target attributes. (a) shows speed, SCR and swerve times of targets. Middle circle shows target numbers w.r.t. target SCR, and inner and outer circles show the swerving times and target speed of different target SCR levels. Numbers in the pie chart represent the proportion of targets at each SCR level. Numbers in the legends of top-right corner and bottom-right corner represent the proportion of targets in each speed level and swerving times. (b) shows the area and shape distribution of target. ``S0-S4" outside the circle represent different target shapes, of which the eccentricity range from $[0,0.2),[0.2,0.4),[0.4,0.6),[0.6,0.8),[0.8,1]$. Lines with different colors represent different target area levels. Radius represents the annotation number, and the area under each color line represents the total annotation number of each target area level.}}\label{fig:Target_attribute}
	\vspace{-.6cm}
\end{figure}

With 200 SWIR images as background images, we totally synthesize 200 video sequences. To avoid data bias and over-fitting, training and test sets are split for 160 and 40 video sequences by the following criterion. 1) Each subset covers all attributes of target (\textit{e.g.,} target number, SCR, moving speed, and swerving times, see section~\ref{sec-Target_Diversity}). 2) Each subset covers all attributes of backgrounds (\textit{e.g.,} background complexity and moving speed, see section~\ref{sec-Complex_Background}). 3) Two subsets are not overlapped. Figure~\ref{fig:Target_attribute} (a) shows instance number and sequence length of sequence number within training and test sets. It can be observed that instance number ranges from 1 to 6. Sequence with 2 instances occupies the largest proportion (\textit{i.e.,} 38\%), and sequence with instance number $\leq$3 is three times more than that of instance number$>$3 (\textit{i.e.,} 77.5\% vs. 22.5\%). In addition, sequence length ranges from 104 to 993. Among them, 26.5\% are short sequences (0-200), 42.5\% are medium sequences (200-600), and 31.0\% are long sequences (600-1000). 

IRSatVideo-LEO offers two types of annotations: 1) Instance-based BBox annotations for IRST detection and tracking. 2) Instance-based mask annotation for IRST detection, semantic \& instance segmentation. 

\subsection{Statistical Properties}

\subsubsection{Rich Target Diversity}\label{sec-Target_Diversity}

We calculate the signal-to-clutter ratio (SCR), speed and swerving times of targets. Among them, SCR is calculated in the local background neighborhood of targets (see Fig.~\ref{fig-Guassian} (b), $d$ is set to $20$ in our paper), and can be formulated as:
\begin{align}\label{eq-SCR}
	SCR=\frac{|\mu_t-\mu_b|}{\sigma_b},
\end{align}
where $\mu_t$ is the mean value of image in target region (\textit{i.e.,} yellow area in Fig.~\ref{fig-Guassian} (b)). $\mu_b$ and $\sigma_b$ are the mean and standard deviation values in the local background region (\textit{i.e.,} purple area in Fig.~\ref{fig-Guassian} (b)). Statistical values in Fig.~\ref{fig:Target_attribute} (a) show that the number of targets with SCR$<$3 is over a third (\textit{i.e.,} 34.8\%), which demonstrates that IRSatVideo-LEO dataset is challenging of not only small (target area$<$$9$$\times$$9$) but also dim targets. In addition, over half of the targets present no swerve and slow speed, while a small number of targets present 1-2 swerving times (46.4\%) and medium or fast speed (38.5\%), enriching the target diversity and increase the difficulty of IRSatVideo-LEO dataset. In conclusion, our dataset contains high-diversity targets with different numbers, SCR, speed and swerving times, and the distribution of these attributes is controllable and in accordance with the real-world scenario. {In addition, Fig.~\ref{fig:Target_attribute} (b) shows the distribution of target area, target shape w.r.t. annotation number. Since targets are simulated by Gaussian kernel, we employ eccentricity $e$ for shape distinguishing. It can be observed that, targets smaller than 10 pixels occupy the largest proportion, and most of the targets are in circular shape.} 

\subsubsection{Complex Background}\label{sec-Complex_Background}
\begin{figure}[t]
	\centering\includegraphics[width=9cm]{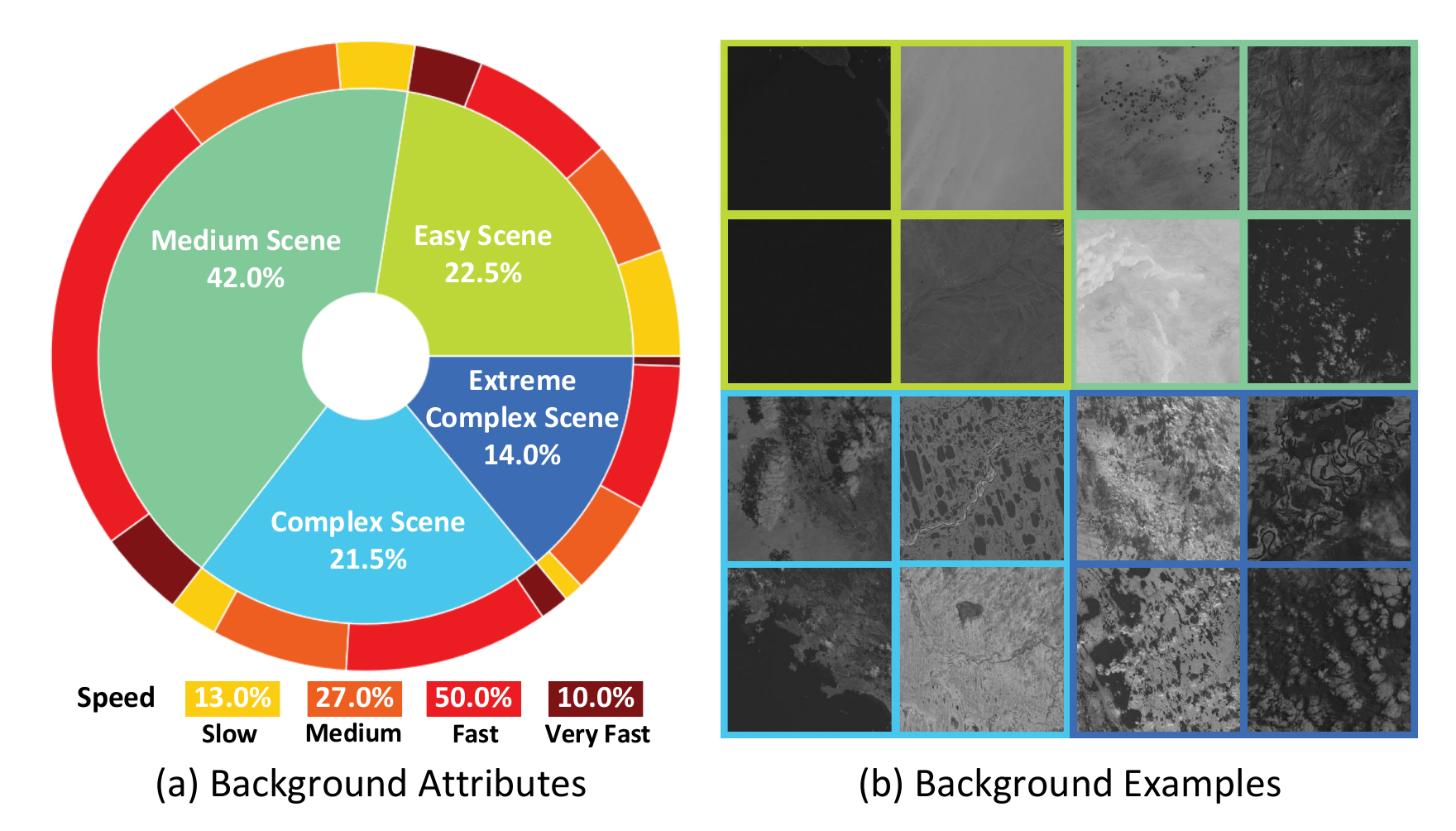}
	\vspace{-.5cm}
	\caption{Illustration of background attributes and example images. (a) shows different attributes of backgrounds. Inner circle shows background complexity w.r.t. sequence numbers, and outer circles show the background moving speed of different background complexity levels. Numbers in the pie chart represent the proportion of sequences in each background complexity. Numbers in the legends represent the proportion of sequences in each background moving speed levels. (b) shows example background images with different complexity levels.}\label{fig-complex_background}
	\vspace{-.2cm}
\end{figure}

Following \cite{Background_complexity}, we evaluate the background complexity of a sequence by the average information entropy and variance of images in the sequence, which can be formulated as:
\begin{align}
	C_t&=-\sum_{s=0}^{255} (s-\bar s(I_t))p_s(I_t) \log(p_s(I_t)),\\
	\bar C&=\sum_{t=1}^{T}C_t/T,
\end{align}
where $C_t$ represents the background complexity of the $t^{th}$ frame $I_t$ in a sequence, and $\bar C$ represents the average one of the whole sequence. $s$ represents the gray level in the histogram of $I_t$, and $p_s(I_t)$ is the probability of $s$ in $I_t$. 
As shown in Fig.~\ref{fig-complex_background} (a), we divide background complexity into four levels: easy $\bar C$$\in$$[0,200)$, medium $\bar C$$\in$$[200,1000)$, complex $\bar C$$\in$$[1000,2000)$ and extreme complex $\bar C$$\in$$[2000,\infty)$. It can be observed that our dataset covers a large range of background complexity (from 9 to 5596). In addition, we visualize some example background images in different levels in Fig.~\ref{fig-complex_background} (b). It can be observed that easy sequences have low background fluctuation, which increases significantly as the level increases. Moreover, we divide background moving speed (\textit{i.e.,} average shift pixels per frame) into four levels: slow$\in$$[0,1/10)$, medium$\in$$[1/10,1/3)$, fast$\in$$[1/3,1)$ and very fast$\in$$[1,2]$. It can be observed that there exists all background moving speed levels in each background complexity level, and over 60\% of sequences are in fast or even higher background moving speed. In conclusion, our dataset contains abundant backgrounds with different complexity and moving speeds.

\subsubsection{{Comparison to Existing IRST Detection Datasets}}
{In this subsection, we make comprehensive comparisons among recent public SIRST datasets \cite{MDvsFA,OSCAR,DNANet,ISNet,NUDT-SIRST-Sea}, MIRST datasets \cite{Hui1,Hui2,SIATD,Anti-UAV,RDIAN} and our IRSatVideo-LEO dataset. As shown in Table~\ref{tab-Statics}, compared to SIRST datasets, MIRST datasets are in much larger scales (\textit{e.g.,} frame number and target number) and focus on targets with lower SCR (temporal information is necessary for MIRST detection). Note that, many MIRST datasets only provide Bounding Box (BBox) or point annotations, which cannot provide a comprehensive performance evaluation among existing MIRST detection methods (\textit{i.e.,} foreground and background segmentation task). Although IRDST and NUDT-MIRSDT datasets provide mask annotations, the land-based imaging system (\textit{e.g.,} small field of view and disparity variations), and PTZ Camera (\textit{e.g.,} irregular background motion results from fast and flexible platform) or static background motion cannot meet the actual conditions of satellite videos (\textit{e.g.,} large field of view without disparity variation with respect to remote sensing imaging system, slow \& regular background motion with respect to satellite cameras). 
In conclusion, IRSatVideo-LEO is the first large-scale MIRST dataset in satellite videos, that exhibits larger data quantity, higher image resolution, LEO satellite-based moving platform, short wave band imaging system, maneuvering small dim targets, groundtruth mask annotations, making it a valuable resource for research and development in satellite-based video surveillance and remote sensing applications.}

\section{Method}\label{Method}

\begin{figure*}[t]
	\centering\includegraphics[width=18cm]{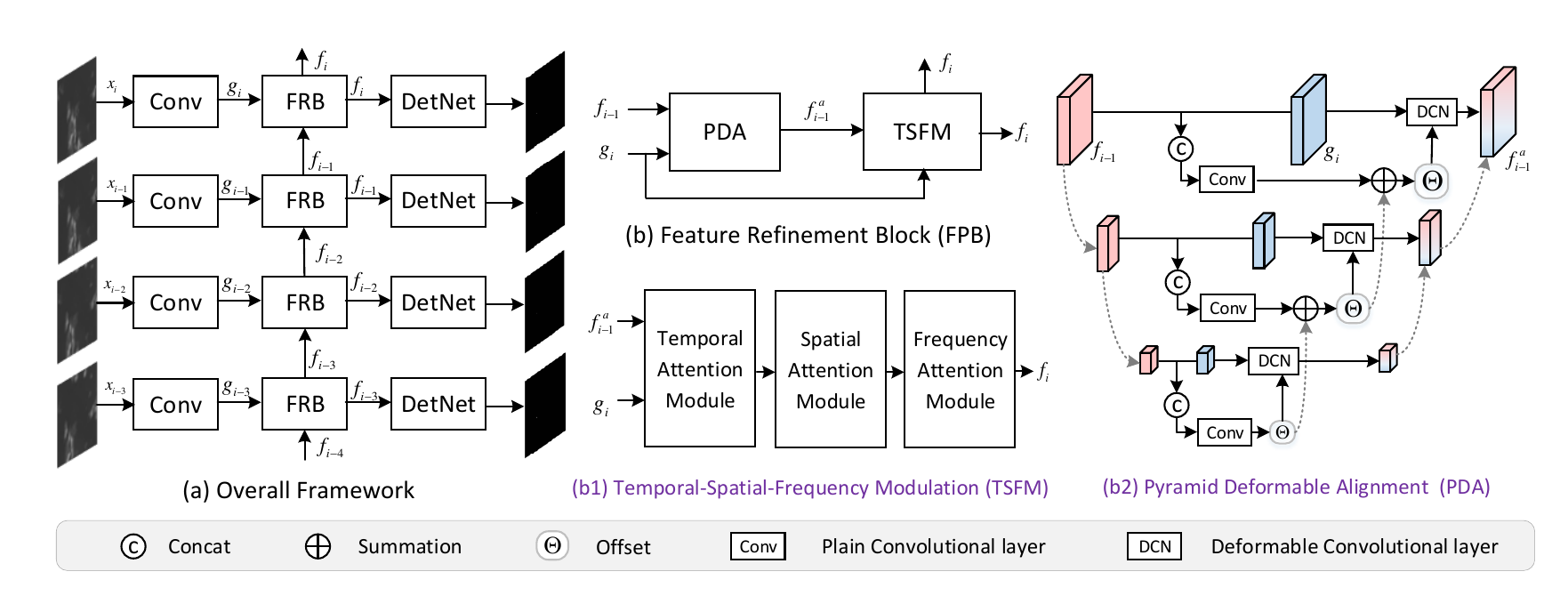}
	\vspace{-.3cm}
	\caption{The proposed architecture. (a) is the overall framework. (b) denotes feature propagation block (FPB), and (b1) and (b2) represent its submodules, including temporal-spatial-frequency modulation (TSFM) module and pyramid deformable alignment (PDA) module.}\label{fig-framework}
	\vspace{-.2cm}
\end{figure*}

\subsection{Overall Framework}
{As shown in Fig.~\ref{fig-framework}, the pipeline of our recurrent feature refinement (RFR) framework can be summarized into three components: 1) 3$\times$3 convolution for initial feature extraction to generate initial feature $g\in \mathbb{R}^{C \times T \times H \times W}$ from the input image sequence $x\in \mathbb{R}^{1 \times T \times H \times W}$. 2) Recurrent feature refinement block (FRB) for recurrent feature alignment, propagation, aggregation and refinement. That is, FRB refines the current feature $g_i\in \mathbb{R}^{C  \times H \times W}$ using recurrently refined previous features $f_{i-1}\in \mathbb{R}^{C  \times H \times W}$, and thus exhibits remarkable efficiency in leveraging long-term temporal dependencies. 3) SIRST detection network for object detection. Note that,  In this paper, we mainly focus on the design of recurrent FRB, which consists of a pyramid deformable alignment (PDA) module for coarse-to-fine motion compensation, and a temporal-spatial-frequency modulation (TSFM) module for multi-dimensional information refinement.} The details of PDA and TSFM are introduced in sections \ref{sec-pda} and \ref{sec-tsfm}.

\subsection{Pyramid Deformable Alignment}\label{sec-pda}
{Given the initial feature sequence $\{g_1, g_2,\cdots, g_T\}$ with temporal length $T$, modulated deformable convolution \cite{DCN,DCNv2} is utilized to recurrently align the previous features $f_{i-1}$ to the current features $g_{i},i\in[2,T]$.} Specifically, for a modulated deformable convolution with kernel size $k_h\times k_w$, the aligned feature $f_{i-1}^{a}$ can be obtained by:
\begin{align}\label{eq-DCN}
	f_{i-1}^{a}(p_0) = \sum\limits_{n = 1}^N {w(p_n) \cdot f_{i-1}(p_0 + p_n+ \Delta p_{n})\cdot \Delta m_{n}},
\end{align}
where $p_0$ denotes a location in the aligned feature. $N$$=$$k_h\times k_w$ represents the total number of the sampling locations, and $p_n$ denotes the $n^{th}$ value in convolutional sampling grid $G$$=$${(-\lfloor\frac{k_h}{2}\rfloor,-\lfloor\frac{k_w}{2}\rfloor),(-\lfloor\frac{k_h}{2}\rfloor,-\lfloor\frac{k_w}{2}\rfloor+1), \cdots,(\lfloor\frac{k_h}{2}\rfloor,\lfloor\frac{k_w}{2}\rfloor)}$. $\lfloor \cdot \rfloor$ is the round down operation. $w(p_n)$ represents the weight in the $n^{th}$ location of convolution kernel. $\Delta p_n$ and $\Delta m_{n}$ represent the corresponding learnable offset and learnable modulation scalar predicted by the concatenation of the previous features and the current features. The process can be defined as:
\begin{align}
	\Delta P_{i-1}, \Delta M_{i-1} =S(\theta( [f_{i-1},g_{i}])), 
\end{align}
where $\Delta P$=$\{\Delta p\}$, $\Delta M$=$\{\Delta m\}$. $\theta(\cdot)$ represents plain convolution layers for feature embedding. $S(\cdot)$ represents channels split operation, and $[\cdot,\cdot]$ represents the concatenation operation. Following \cite{DCN,DCNv2}, we employ bilinear interpolation to generate exact values since $p_0 + p_n+ \Delta p_{n}$ is generally fractional. {Compared with optical flow-based methods, deformable convolution can perform implicit motion compensation to alleviate target energy loss and outline blur caused by wrapping.}

{Considering the scale differences between background motion and target motion, we use a pyramid structure to decompose the motion, and perform motion compensation in a coarse-to-fine manner.} Specifically, we first employ strided convolution filters (\textit{i.e.,} gray down arrow with dash lines in Fig.\ref{fig-framework} (b2)) to generate L-level feature pyramids $\{f_{i-1}^{l}\}$ and $\{g_{i}^{l}\}$, $l\in[1,L]$ by a scale factor of 2. L is set to 3 in our paper. Within each feature pyramid level, deformable convolutions are used for feature alignment. Note that, except the lowest level L, offsets and aligned features in $l^{th}$ level are generated by the predicted ones together with the $\times$2 upsampled ones (\textit{i.e.,} gray up-arrow with dash lines in Fig.\ref{fig-framework} (b2)) from the upper $(l+1)^{th}$ level. The process can be defined as:
\begin{align}
	\Delta P_{i-1}^{l}&=\theta( [f_{i-1},g_{i}], (\Delta P_{i-1}^{l+1})^{\uparrow 2}), \\
	(f_{i-1}^a)^{l} &= \theta(\rm{DCN}(f_{i-1}^{l},\Delta P_{i-1}^{l}), ((f_{i-1}^a)^{l+1})^{\uparrow 2}),
\end{align}
where $(\cdot)^{\uparrow 2}$ represents bilinear upsample interpolation by a scale factor of 2. $\rm{DCN}$ represents modulated deformable convolution illustrated in Eq.~\ref{eq-DCN}. {Note that, in the higher level of feature pyramid, we mainly focus on large-scale target motion. Subsequently, we propagate the offsets and aligned features to lower level to achieve more accurate background motion compensation. Through such a coarse-to-fine manner, sub-pixel accuracy can be obtained for precise motion compensation and target information refinement. In addition, PDA serves as a feature alignment module, and is also jointly optimized with the detection network for end-to-end training without additional supervision \cite{Zhu2023a} or other pretrained models \cite{Luo2022}.}

\begin{table*}
	\begin{center}
		\centering
		\renewcommand\arraystretch{1.4}
		\caption{{\scriptsize{Parameter settings of 4DISTD \cite{4DISTD}, ASTTV\_NTLA \cite{ASTTV-NTLA}, MSLSTIPT \cite{MSLSTIPT}, SRSTT \cite{SRSTT}, IMNN-LWEC\cite{IMNN-LWEC}, NFTDGSTV \cite{NFTDGSTV}, and RCTVW \cite{RCTVW}}}}\label{tab-parameters}
		\scriptsize \vspace{-.1in}
		{\begin{tabular}{lc}
				\toprule[1pt]
				Method&Parameters\\
				\midrule
				4DISTD \cite{4DISTD}&Sliding step: 70, Patch size: $ 70 \times 70 $, $L$=$15$, $\lambda_1$=$0$, $\lambda_2$=$100$, $\delta_n$=$1/L$, $\beta_n$=$1e-4$, $\gamma$=$1e-4$, $\tau$=$1e-7$, $\epsilon$=$1e-3 $\\
				ASTTV\_NTLA \cite{ASTTV-NTLA}&$ L $=$ 3 $, $ {\lambda _{tv}} $=$ 0.005 $, $ {\lambda _s} $=$ \frac{H}{{\sqrt {\max \left( {M,N} \right) * L} }} $,$ H $=$ 8 $, $ {\lambda _3} $=$ 100 $ \\
				MSLSTIPT \cite{MSLSTIPT}&$ \lambda $=$ {1 \mathord{\left/
						{\vphantom {1 {\sqrt {{n_3}\max ({n_1},{n_2})} }}} \right.
						\kern-\nulldelimiterspace} {\sqrt {{n_3}\max ({n_1},{n_2})} }} $,$ L $=$ 6 $, Patch size: $ 30 \times 30 $, $ p $=$ 0.8 $ \\
				SRSTT \cite{SRSTT}&$\lambda_1$=$0.05$, $\lambda_2$=$0.1$, $\lambda_3$=$100$, $\mu$=$0.01$, $\rho$=$1.3$, $\tau$=$1e-7$\\
				IMNN-LWEC\cite{IMNN-LWEC}&Sliding step: 15, Patch size: 15$\times$15, L$=$3, $ {\lambda _1} $=$ \frac{{{\lambda _L}}}{{\sqrt {\max \left( {{n_1},{n_2}} \right) * {n_3}} }} $, $w:5$, $ {\lambda _L} $=$ 1.4 $, $ {\lambda _2} $=$ {\lambda _3} $=$ 50{\lambda _1} $ , $ \varepsilon$=$10^{ - 4} $ , $ {k} $=$ 1.5 $, $\rho $=$ \frac{1}{w} $=$ {\beta _{{k_1}{k_2}}} $\\
				NFTDGSTV \cite{NFTDGSTV}&$ L $=$ 3 $, $ H $=$ 4 $, $ {\lambda _1} $=$ 0.01 $, $ {\lambda _2} $=$ \frac{H}{{\sqrt {\max \left( {M,N} \right) * L} }} $, $ {\lambda _s} $=$ 0.001 $\\
				RCTVW \cite{RCTVW}&$ H $=$ 4 $, $ \lambda $=$ \frac{H}{{\sqrt {\max \left( {m,n} \right) \times L} }}$, $ \alpha $=$ 0.8$, $ L $=$ 8 $, $ r$=$6 $, $\beta $=$ 2e - 4 $, $ \rho $=$ 1.1 $, $\mu $=$ 0.01$ \\
				\bottomrule[1pt]
		\end{tabular}}
	\end{center}
\end{table*}

\subsection{Temporal-Spatial-Frequency Modulation}\label{sec-tsfm}

{Temporal information aggregation and target information preservation and enhancement are critical for recurrent framework. The reasons can be summarized as: 1) errors (\textit{e.g.,} misalignment, occlusion, blurry regions, etc) are progressively accumulated and amplified through recurrent feature propagation, which significantly affect the subsequent detection performance. 2) Targets are small and feature-scarce, which are easily immersed and lost in recurrent feature propagation. Therefore, dynamic feature fusion of adjacent frames and target feature enhancement is crucial to ensuring both the effectiveness and efficiency of the recurrent feature propagation process. To handle the aforementioned problem, we propose a temporal-spatial-frequency modulation module to allocate per-pixel weight for temporal information aggregation and spatial \& frequent target information enhancement. Specifically, we subsequently employ temporal attention, spatial attention and frequent attention to fully exploit the temporal consistency, spatial saliency \& frequent distinguishability of infrared small target, as shown in Fig.~\ref{fig-framework} (b1).}

{Temporal attention \cite{STDMANet} aims to eliminate the misaligned, occluded, and blur pixels of aligned features for temporal feature aggregation.} Specifically, we first calculate the feature similarity between the aligned feature $f_{i-1}^a$ and the current feature $g_{i}$ in an embedding space to generate the temporal attention map $M_{i}^s$. The process can be defined as:
\begin{align}
M_{i}^s = \mathcal{P}_{Sigmoid}(\theta(f_{i-1}^a)^{T}\theta(g_{i})),
\end{align}
where $\mathcal{P}_{Sigmoid}$ represents the Sigmoid function that serves as an activator to normalize the temporal attention and stabilize the training process. Awaring of the pixel-level errors, the aligned feature is then multiplied by attention map for temporal information modulation. A plain convolutional layer is also adopted for temporal feature fusion. The process can be defined as:
\begin{align}
f_i^s = \theta([f_{i-1}^a\odot M_{i}^s, g_i]),
\end{align}
where $\odot$ is the Hadamard production.

\begin{table*}
	\footnotesize
	\renewcommand\arraystretch{1.3}
	\centering
	\caption{{$P_d$($\times$10$^{-2}$), $F_a$($\times$10$^{-6}$) values achieved by different methods on IRSatVideo-LEO dataset. ``\#Params.'' represents the number of parameters. FLOPs is computed based on input image sequence with a resolution of 20$\times$256$\times$256, and time is computed based on input image sequence with a resolution of 20$\times$1024$\times$1024. The best results are shown in red and the second best results are shown in blue.}}\label{tab-comp}
	\setlength{\tabcolsep}{0.8mm}
	\begin{tabular}{|l|c|c|c|c|ccc|ccc|ccc|ccc|}
		\hline
		&{\multirow{2}*{Methods}}&\multirow{2}*{\#Params.}&\multirow{2}*{FLOPs}&\multirow{2}*{Time}&\multicolumn{3}{c|}{Easy}&\multicolumn{3}{c|}{Medium }&\multicolumn{3}{c|}{Complex}&\multicolumn{3}{c|}{Total}\\\cline{6-17}
		&&&&&$P_d$$\uparrow$&$F_a$$\downarrow$&$AUC$$\uparrow$&$P_d$$\uparrow$&$F_a$$\downarrow$&$AUC$$\uparrow$&$P_d$$\uparrow$&$F_a$$\downarrow$&$AUC$$\uparrow$&$P_d$$\uparrow$&$F_a$$\downarrow$&$AUC$$\uparrow$\\\hline
		\multirow{10}*{\rotatebox{90}{SIRST Detection}}&ACM \cite{ACM}&0.40M&\textcolor{red}{7.40G}&\textcolor{blue}{0.10} &94.64 &10.54 &94.61 &74.28 &29.47 &73.93 &65.81 &32.46 &67.89 &79.18 &23.69 &79.32 \\
		&ALCNet \cite{ALCNet}&0.43M &\textcolor{blue}{7.50G}&\textcolor{red}{0.09} &96.50 &9.42 &96.69 &73.80 &19.03 &74.59 &65.73 &19.05 &66.22 &79.56 &15.74 &79.91 \\
		&DNA-Net \cite{DNANet}&4.69M&283.59G&6.54 &79.77 &1.03 &80.67 &46.68 &5.67 &46.29 &61.50 &8.30 &64.40 &60.83 &4.70 &61.72 \\
		&ISNet \cite{ISNet}&0.97M&610.08G&1.81 &96.99 &17.96 &97.00 &72.49 &14.63 &72.84 &63.87 &15.39 &64.23 &78.72 &15.95 &78.78 \\
		&UIU-Net \cite{UIU-Net}&50.54M&1087.52G&1.98 &94.78 &4.11 &94.70 &65.58 &9.56 &65.59 &62.23 &7.08 &62.29 &74.51 &7.11 &74.37 \\
		&RDIAN \cite{RDIAN}&\textcolor{red}{0.22M}&74.15G&1.23 &95.01 &38.45 &95.38 &75.46 &48.65 &76.37 &66.01 &20.75 &66.76 &79.88 &38.59 &80.39 \\
		&ISTDU-Net \cite{ISTDU-Net}&2.75M&156.75G&1.41 &97.87 &12.20 &97.91 &76.37 &19.44 &76.81 &67.72 &16.90 &70.11 &81.60 &16.36 &82.19 \\
		&ResUNet \cite{ResUNet}&\textcolor{blue}{0.91M}&76.33G&0.37 &96.32 &20.98 &96.82 &72.66 &20.72 &73.97 &65.41 &14.47 &68.73 &78.91 &19.34 &80.25 \\
		&RPCANet \cite{RPCANet}&0.68M&891.15G&3.46 &97.49 &23.56 &97.40 &71.73 &20.71 &71.77 &65.71 &17.22 &65.74 &78.94 &20.87 &78.78 \\
		&AGPCNet \cite{AGPCNet}&12.43M&863.32G&1.73 &97.71 &9.25 &97.73 &72.94 &14.54 &73.05 &65.44 &15.01 &67.86 &79.51 &12.84 &79.95 \\
		\hline
		\multirow{12}*{\rotatebox{90}{MIRST Detection}}&MSLSTIPT \cite{MSLSTIPT}&-&-&593.30 &1.37 &\textcolor{red}{0.17} &97.82 &0.83 &\textcolor{red}{0.01} &82.70 &0.00 &\textcolor{red}{0.01} &38.18 &0.82 &\textcolor{red}{0.06} &73.52 \\
		&NFTDGSTV \cite{NFTDGSTV}&-&-&1019.17 &19.68 &\textcolor{blue}{1.02} &98.33 &18.20 &203.03 &83.15 &28.30 &131.36 &81.58 &21.01 &115.78 &86.98 \\
		&RCTVW \cite{RCTVW}&-&-&6.73 &20.63 &1.37 &59.15 &23.36 &1319.96 &26.16 &29.17 &4486.81 &36.87 &23.80 &1732.64 &39.38 \\
		&IMNN-LWEC \cite{IMNN-LWEC}&-&-&3203.45 &76.15 &63.65 &91.05 &38.69 &586.64 &57.92 &48.41 &916.44 &65.97 &53.20 &500.01 &70.64 \\
		&SRSTT \cite{SRSTT}&-&-&4693.61 &72.01 &516.64 &96.01 &46.73 &7.07e4 &58.34 &39.97 &2.40e5 &44.63 &53.46 &9.28e4 &65.20 \\
		&ASTTV-NTLA \cite{ASTTV-NTLA}&-&-&1148.70 &88.83 &8.54 &90.50 &63.13 &106.06 &66.83 &73.92 &71.81 &79.50 &73.96 &64.58 &77.50 \\
		&4DISTD\cite{4DISTD}&-&-&3536.69 &85.03 &12.49 &96.62 &61.54 &9.62 &89.09 &56.95 &27.92 &82.92 &68.31 &14.91 &92.48 \\
		\cline{2-17}
		&STDMANet \cite{STDMANet}&11.88M&62.98G&19.72&98.04&4.90&98.06&86.86&4.63&87.77&82.43&\textcolor{blue}{2.35}&{83.31}&89.96&\textcolor{blue}{4.10}&90.13 \\
		&DNANet\_DTUM \cite{DTUM}&1.21M&176.66G&23.45 &97.65 &7.60 &97.87 &82.80 &23.46 &85.53 &78.96 &5.02 &83.12 &86.88 &13.68 &89.02 \\
		&ResUNet\_DTUM \cite{DTUM}&0.30M&40.88G&3.09 &99.11 &33.82 &99.16 &\textcolor{blue}{87.93} &76.78 &89.20 &81.32 &12.94 &83.74 &90.19 &47.03 &91.65 \\
		&UIU-Net\_DTUM \cite{DTUM}&51.04M&1151.19G&11.58 &98.81 &56.94 &98.77 &87.77 &89.69 &\textcolor{blue}{88.92} &69.70 &1107.83 &70.10 &87.51 &318.10 &87.82 \\
		&ACM\_RFR&0.50M&79.88G&1.63&96.32&55.86&96.79&76.51&68.92&72.88&71.57&51.87&68.41&82.12&59.92&80.67\\
		&ALCNet\_RFR&0.53M&79.39G&1.61 &97.16 &13.58 &97.45 &77.11 &7.91 &78.85 &70.42 &{4.70} &72.13 &82.29 &9.10 &83.40 \\
		&DNA-Net\_RFR&4.80M&358.16G&8.05 &97.77 &6.21 &97.97 &84.13 &\textcolor{blue}{4.15} &84.60 &81.67 &{3.95} &81.90 &88.11 &4.81 &88.36 \\
		&ISTUD-Net\_RFR&2.86M&231.41G&2.89 &\textcolor{blue}{99.22} &21.22 &\textcolor{blue}{99.23} &87.23 &24.36 &87.91 &\textcolor{blue}{83.14} &15.87 &\textcolor{blue}{83.80} &\textcolor{blue}{90.31} &21.29 &\textcolor{blue}{90.68} \\
		&ResUNet\_RFR&1.01M&150.90G&1.91 &\textcolor{red}{99.49} &23.14 &\textcolor{red}{99.38} &\textcolor{red}{89.27} &19.62 &\textcolor{red}{89.51}&\textcolor{red}{84.33} &10.07 &\textcolor{red}{84.48} &\textcolor{red}{91.58} &18.58 &\textcolor{red}{91.59} \\
		\hline
	\end{tabular}
\vspace{-.3cm}
\end{table*}

{Spatial attention \cite{DNANet} aims to exploit the spatial saliency prior of small target for adaptive feature enhancement.} Specifically, spatial attention map is generated by the pixel-level spatial maximum and average information in an embedding space, which is then multiplied by the input feature for spatial information modulation. The process can be defined as:
\begin{align}
	M_{i}^{t} &= \mathcal{P}_{Sigmoid}(\theta[\mathcal{P}_{max}(f_i^s),\mathcal{P}_{avg }(f_i^s)]),\\
	f_i^{st} &= f_i^s\odot M_{i}^{t},
\end{align}
where $M_{i}^{t}$ represents the spatial attention map. $f_i^{s}$ and $f_i^{st}$ represent the output feature of temporal and spatial information modulation. $\mathcal{P}_{max}$ and $\mathcal{P}_{avg}$ represent max and average operation along channel dimension, respectively. 

{Frequent attention \cite{FcaNet} aims to exploit the frequent distinguishability prior of small target for enriched representation.} Specifically, input feature $f_i^{st}\in \mathbb{R}^{C \times H \times W}$ is first decomposed by channels to several parts $X^i \in \mathbb{R}^{C^{\prime} \times H \times W}, i \in\{0,1, \cdots, n-1\}$ with corresponding frequency components by 2D discrete cosine transform (DCT). $C^{\prime}=C/n$. The process can be defined as:
\begin{equation}
\begin{aligned}
		 \text {Freq}^i&= \mathrm{DCT}^{u_i, v_i}\left(X^i\right)\\
		 &=\sum_{h=0}^{H-1} \sum_{w=0}^{W-1} X_{:, h, w}^i B_{h, w}^{u_i, v_i} \\
		& \text { s.t. } , i \in\{0,1, \cdots, n-1\},\\
\end{aligned}
\end{equation}
where $\text {Freq}^i\in \mathbb{R}^{C^{\prime}}$ is the frequency sub-vector, and $u_i, v_i$ are the 2D indices of frequency components corresponding to $X^i$. Then, all frequency sub-vectors are concatenated to obtain the frequency vector. Afterwards, the vector is then mapped to an embedding space and activated by Sigmoid function to generate the frequent attention map. Finally, the output feature is generated by multiplication between the input feature and the frequent attention map for frequent information modulation. The process can be defined as:
\begin{align}
	M_{i}^f &= \mathcal{P}_{Sigmoid}(\theta([\text {Freq}^0,\text {Freq}^1,\cdots,\text {Freq}^{n-1}])),\\
	f_i^{stf} &= f_i^{st}\odot M_{i}^{f},
\end{align}
where $M_{i}^f$ represents the frequent attention map, and $f_i^{stf}$ represents the temporal-spatial-frequent modulated feature. Note that, we employ the two-step criterion \cite{FcaNet} to select the top-k significant frequency components.

\section{Experiments}\label{experiments}
In this section, we first introduce the experiment settings, including evaluation metrics and implemental details. Then, we apply our method to the state-of-the-art SIRST detection methods, and make comparisons with several state-of-the-art SIRST and MIRST detection methods. Finally, we present ablation studies to validate our design choice. 

\subsection{Experimental Settings}

\subsubsection{Evaluation Metrics}
{In this paper, we employ the probability of detection ($P_d$), false alarm rate ($F_a$), receiver operating characteristic curve (ROC), and area under curve (AUC) to evaluate the detection performance. }

\textit{Probability of Detection:} {$P_d$ is used to evaluate the localization capability of detection algorithms, and can be defined as:}
\begin{align}
P_d=\frac{{{\rm{TD}}}}{{{\rm{AT}}}},
\end{align}
where $\rm{TD}$ and $\rm{AT}$ are the number of truly detected targets and all targets, respectively. Targets are classified as truly detected ones when their centroid deviation is lower than a pre-defined deviation threshold. Following \cite{DNANet,LESPS}, we set the threshold to 3 in our paper.

\textit{False Alarm Rate:} {$F_a$ is used to evaluate the false alarm suppression capability of detection algorithms, and can be defined as:}
\begin{align}
	F_a=\frac{{{\rm{FD}}}}{{{\rm{NP}}}},
\end{align}
where $\rm{FD}$ and $\rm{NP}$ are the number of false detected pixels and all image pixels, respectively. Pixels are classified as falsely detected ones, when their centroid deviation is larger than the pre-defined deviation threshold. Following \cite{DNANet,LESPS}, we set the threshold to 3 in our paper.

\begin{figure*}[t]
	\centering\includegraphics[width=18cm]{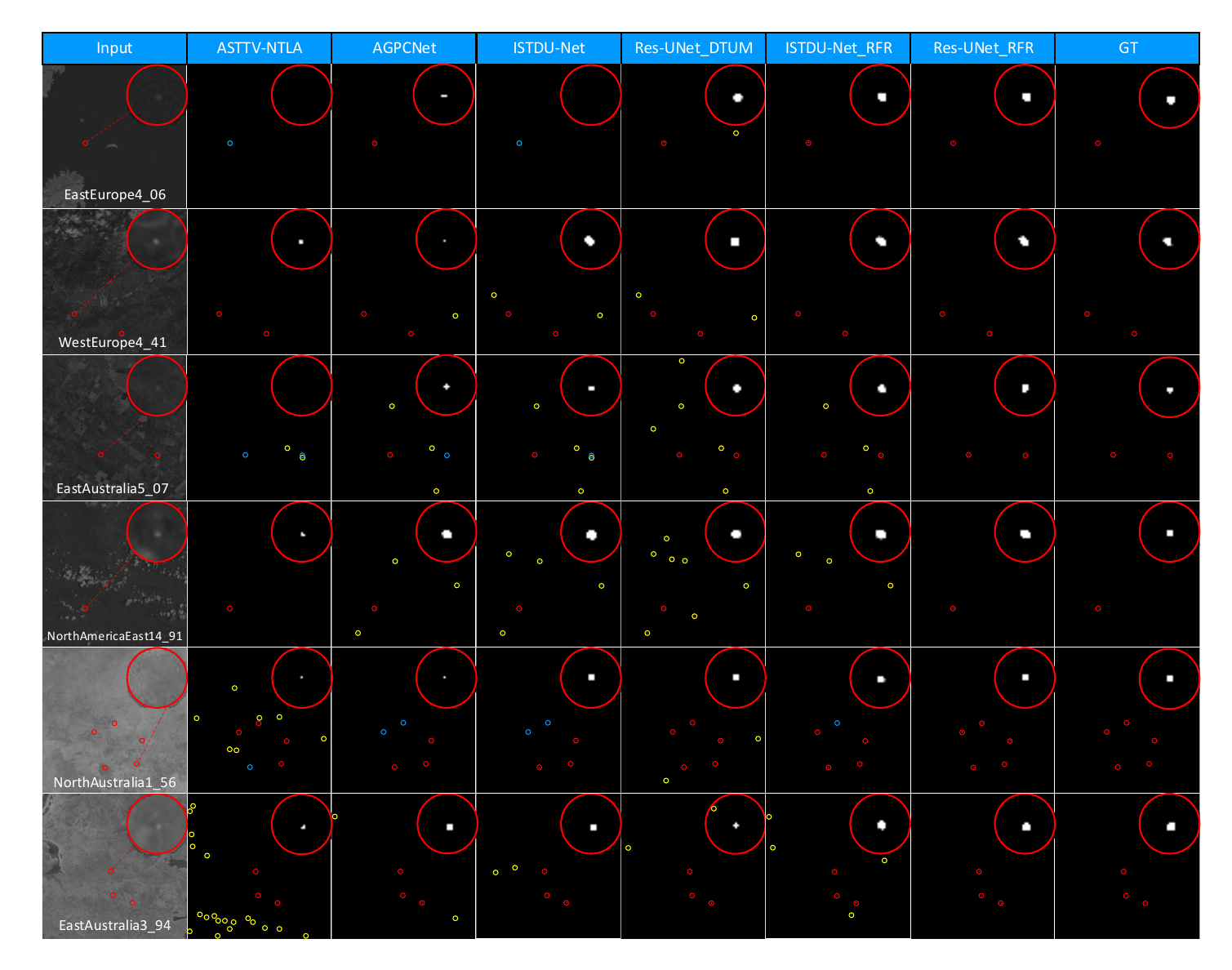}
	\vspace{-.3cm}
	\caption{Qualitative results of different methods. For better visualization, we show the zoom-in target regions on the top-right, which are highlighted by red circles. Missing detection and false alarms are highlighted by blue and yellow circles.}\label{fig-visual_comp}
	\vspace{-.2cm}
\end{figure*}

{\textit{Receiver Operation Characteristics:} ROC \cite{DNANet} is used to describe the trends of $P_d$ under varying $F_a$.}

{\textit{Area Under Curve:} AUC \cite{DTUM} is the area under ROC, and larger value represents better detection performance.}

\subsubsection{Implementation Details}\label{sec_imp}

During the training phase, we randomly selected $K$ consecutive frames from all video clips in the training set of IRSatVideo-LEO, and randomly cropped a $K\times128\times128$ sequence patch as the input. We followed \cite{DNANet,LESPS,DTUM} to augment the training data by random flipping and rotation. Note that, we control the ratio of positive samples to 0.9 to accelerate the training process. Different from the widely used short temporal sliding window framework (mainly less than 7 input frames), our methods employ much longer sequence input (\textit{i.e.,} $K=10, 20, 40$ are investigated in section~\ref{sec-abl-RF}) for joint optimization. For test, the whole sequence is input into our recurrent feature refinement framework to capture long-term temporal dependency.

Our network was trained using the Soft-IoU loss function \cite{DNANet,DTUM} and optimized by the Adam method \cite{Adam} with $\lambda_1$ = 0.9, $\lambda_2$ = 0.999. 
Batch size was set to 3. The learning rate was initially set to $5e$-$4$ and halved every 5 epochs. We trained our network from scratch for 20 epochs. All experiments were implemented in PyTorch on a PC with an Nvidia GeForce RTX 3090 GPU. 

\subsection{Comparison to the State-of-the-art Methods}

{In this subsection, we apply our recurrent feature refinement framework with 5 existing state-of-the-art SIRST detection methods (\textit{i.e.,} ACM \cite{ACM}, ALCNet \cite{ALCNet}, DNA-Net \cite{DNANet}, ISTDU-Net \cite{ISTDU-Net} and ResUNet \cite{ResUNet}), and make comparisons to the state-of-the-art SIRST detection methods and MIRST detection methods. For SIRST detection methods, we adopt 10 recent state-of-the-art data-driven methods (\textit{i.e.,} ACM \cite{ACM}, ALCNet \cite{ALCNet}, DNA-Net \cite{DNANet}, ISNet \cite{ISNet}, UIU-Net \cite{UIU-Net}, RDIAN \cite{RDIAN}, ISTDU-Net \cite{ISTDU-Net}, ResUNet \cite{ResUNet}, RPCANet \cite{RPCANet}, AGPCNet \cite{AGPCNet}). For MIRST detection methods, we adopt 7 model-driven methods (\textit{i.e.,} 4DISTD \cite{4DISTD}, ASTTV\_NTLA \cite{ASTTV-NTLA}, MSLSTIPT \cite{MSLSTIPT}, SRSTT \cite{SRSTT}, IMNN-LWEC\cite{IMNN-LWEC}, NFTDGSTV \cite{NFTDGSTV}, and RCTVW \cite{RCTVW}) and 4 state-of-the-art data-driven methods (\textit{i.e.,} STDMANet \cite{STDMANet}, DNANet+DTUM \cite{DTUM}, ISTDUNet+DTUM \cite{DTUM} and ResUNet+DTUM \cite{DTUM}).} The parameters of 7 model-driven MIRST methods are shown in Table~\ref{tab-parameters}. For fair comparison, we retrain all the compared data-driven methods on IRSatVideo-LEO dataset from scratch under the same loss function and implemental details. Note that, the training details of SIRST detection methods are inherited from the BasicIRSTD toolbox \cite{BasicIRSTD}, and the training details of data-driven MIRST detection methods are the same as those in section~\ref{sec_imp}. Note that, we adopt a fixed threshold of 0.5 for all the compared algorithms.

\begin{figure*}[t]
	\centering\includegraphics[width=18cm]{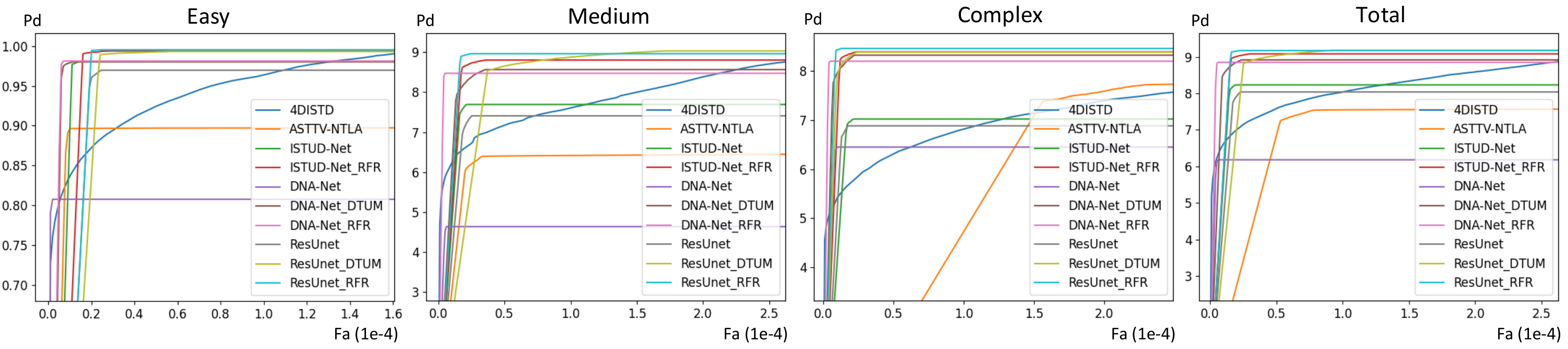}
	\vspace{-.3cm}
	\caption{ROC curves of different methods on IRSatVideo-LEO with easy scene (a), medium scene (b), complex scene (c), and all test set (d).}\label{fig-visual_roc}
	\vspace{-.2cm}
\end{figure*}

\subsubsection{Quantitative Results}

We test different methods under three types of scenes for a more comprehensive performance evaluation. a) \textit{Easy scene} represents the scene of target SCR higher than $6$ and background complexity lower than $1000$. The total number of easy scenes is 18. b) \textit{Medium scene} represents the scene of target SCR and background complexity both higher or lower than $6$ and $1000$. The total number of medium scenes is 13. c) \textit{Complex scene} represents the scene of target SCR lower than $6$ and background complexity higher than $1000$. The total number of complex scenes is 9. 

\begin{table}
	\footnotesize
	\renewcommand\arraystretch{1.2}
	\centering
	\caption{\footnotesize{{$P_d$($\times 10^2$) and $F_a$($\times 10^6$) values achieved by main variants of recurrent framework on IRSatVideo-LEO dataset. ``-TSW$_t$" represents temporal sliding window framework with temporal length $t$ and stride 1. ``-R$_{t-t}$" represents recurrent framework with temporal length $t$ for training, and temporal sliding window framework for test with temporal length $t$ and stride $t$. ``-R$_{t-seq}$" represents recurrent framework with temporal length $t$ for training, and recurrent framework for test with the entire image sequence. ``\#Params.'' represents the number of parameters. FLOPs is computed based on input image sequence with spatial resolution of 256$\times$256 and temporal length of $t$ for ``-TSW", $20$ for ``-R". $^*$ represents the final model. Best results are shown in boldface.}}}\label{tab-Different_Frameworks}
	\begin{tabular}{|l|c|c|c|c|}
		\hline
		Frameworks&\#Params.$\downarrow$&FLOPs$\downarrow$&$P_d$$\uparrow$&$F_a$$\downarrow$\\\hline
		ResUNet&0.905M&76.325G&78.914 &19.339 \\
		ResUNet-TSW$_5$&1.018M&45.200G&81.410 &55.497 \\
		ResUNet-TSW$_7$&1.018M&33.956G&83.558 &49.237 \\
		ResUNet-TSW$_{20}$&1.018M&11.441G&75.336 &17.558 \\
		ResUNet-R$_{20-20}$&1.012M &150.895G&90.525 &13.287 \\
		ResUNet-R$_{10-seq}$&1.012M &150.895G&91.025 &24.802 \\
		ResUNet-R$_{20-seq}$$(*)$&1.012M &150.895G&91.455 &10.760 \\
		ResUNet-R$_{40-seq}$&1.012M &150.895G&91.579 &18.579 \\
		\hline
	\end{tabular}
\end{table}

{Quantitative results are shown in Table~\ref{tab-comp}. Compared with model-driven MIRST detection methods, data-driven methods introduce over 20 $P_d$ and AUC improvements and significant suppressed false alarms, indicating the powerful modeling capability of deep learning-based methods. 
Compared with SIRST detection methods, MIRST detection methods introduce over 10 $P_d$ and AUC improvements on average due to fully exploited additional temporal information. Among MIRST detection methods, ResUNet\_RFR achieves the highest $P_d$ and AUC score with reasonable false alarm increases. Note that, performance decreases as the background complexity increases and target SNR decreases. However, our methods show high robustness to complex scenes with higher $P_d$, AUC scores and lower $F_a$ scores. ROC curve results are shown in Fig.~\ref{fig-visual_roc}. It can be observed that $P_d$ of ResUNet-RFR reaches 1 faster than other compared methods, which also demonstrates the effectiveness and superiority of our method. Moreover, our RFR framework can be equipped with different backbones to achieve substantial performance improvements, which fully demonstrate the generalization of our method.}

\subsubsection{Qualitative Results}

Qualitative results are shown in Fig.~\ref{fig-visual_comp}. It can be observed that our method outperforms all compared approaches to achieve remarkable qualitative results by accurate target localization and shape segmentation with minimal false alarms. Within the compared methods, traditional approaches present constrained capabilities in handling complex backgrounds, and often exhibit a propensity to generate a considerable number of false alarms (\textit{e.g.,} NorthAustralia1\_56 and EastAustralia3\_94). In addition, ResUNet\_DTUM tends to misidentify moving background clutters as targets, resulting in a substantial number of false alarms (\textit{e.g.,} EastAustralia5\_07 and NorthAmericaEast14\_91). Compared with ResUNet\_DTUM, ResUNet\_RFR is more robust to satellite motion due to unified alignment-detection design. Moreover, as shown in lines 6-7 of Fig.~\ref{fig-visual_comp}, our RFR framework can be equipped with diverse SISRT detection techniques to achieve better performance.

\subsubsection{{Computational Efficiency}}

{The computational efficiency (the number of parameters, FLOPs, and running time) are evaluated in Table~\ref{tab-comp}. Note that, running time is the total time tested on a 20$\times$1024$\times$1024 image sequence with an Intel(R) Core(TM) i7-8700 CPU @ 3.20GHz, and is averaged over 100 runs. As compared with model-driven methods, data-driven methods achieve large-margin improvements in both detection performance and running time. Among all data-driven methods, MIRST detection methods show superior detection performance with reasonable increase in computational cost and running time.}

\begin{table}
	\footnotesize
	\setlength{\tabcolsep}{2.9mm}
	\renewcommand\arraystretch{1.2}
	\centering
	\caption{\footnotesize{{$P_d$($\times 10^2$) and $F_a$($\times 10^6$) values achieved by main variants of pyramid deformable alignment on IRSatVideo-LEO dataset. ``-Flow" represents optical flow-based method for feature alignment. ``-Att" represents cross attention module for feature alignment. ``-DA" represents deformable convolution for feature alignment. ``-PDA" represents pyramid deformable convolution for feature alignment. ``\#Params.'' represents the number of parameters. FLOPs is computed based on input image sequence with a resolution of 20$\times$256$\times$256. $(*)$ represents the final model. Best results are shown in boldface.}}}\label{tab-PDA}
	\begin{tabular}{|l|c|c|c|c|}
		\hline
		Frameworks&\#Params.$\downarrow$&FLOPs$\downarrow$&$P_d$$\uparrow$&$F_a$$\downarrow$\\\hline
		ResUNet w/o&0.917M&92.480G&83.149 &64.886 \\
		ResUNet-Flow&0.971M&124.608G&85.715 &48.197 \\
		ResUNet-Att&0.937M&105.966G&84.252 &29.544 \\
		ResUNet-DA&0.939M&122.107G&90.726 &13.284 \\
		ResUNet-PDA$(*)$&1.012M &150.895G&91.579 &18.579 \\
		\hline
	\end{tabular}
\end{table}

\begin{figure}[t]
	\centering\includegraphics[width=9cm]{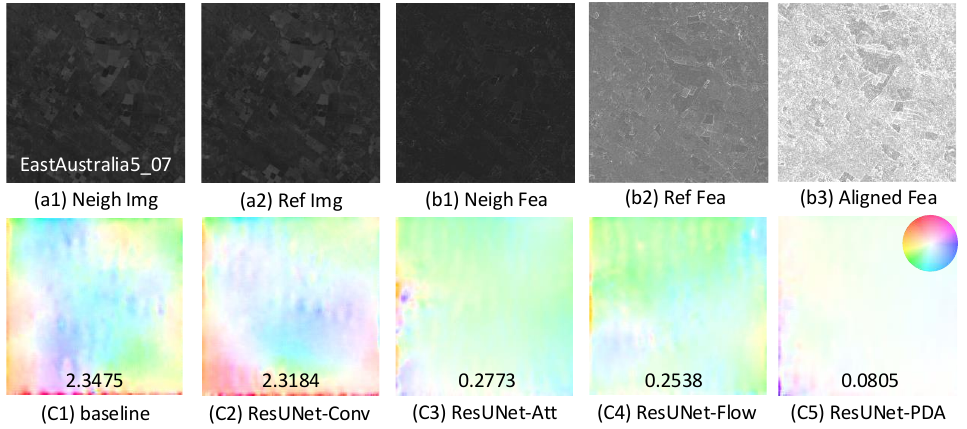}
	\vspace{-.3cm}
	\caption{Visual results of different alignment methods. (a1), (a2) show the neighborhood and reference images and (b1), (b2) show the corresponding features. (b3) shows the aligned features generated by different alignment methods, and (c2)-(c5) shows the flow (derived by RAFT \cite{RAFT}) between aligned and reference features of ResUNet w/o. (c1) shows the flow between neighborhood and reference features as the baseline results. Flow field color coding scheme is shown on the right. The direction and magnitude of the displacement vector are represented by hue and color intensity, respectively. Flow magnitudes are also presented within each flow.}\label{fig-visual_align}
	\vspace{-.2cm}
\end{figure}

\subsection{Ablation Study}
In this subsection, we conduct ablation experiments on our RFR framework with several variants to investigate the potential benefits introduced by our proposed modules and design choices. If not specific, ResUNet \cite{ResUNet} is used as the baseline SIRST detection network. 

\subsubsection{Recurrent Framework}\label{sec-abl-RF}

\begin{table}
	\footnotesize
	\renewcommand\arraystretch{1.2}
	\centering
	\caption{\footnotesize{{$P_d$($\times 10^2$) and $F_a$($\times 10^6$) values achieved by main variants of temporal-spatial-frequent modulation on IRSatVideo-LEO dataset. ``-TM" represents temporal attention for feature modulation. ``-TSM" represents cascaded temporal attention and spatial attention for feature modulation. 
				``-TSFM" represents cascaded temporal attention, spatial attention, and frequency attention for feature modulation. ``\#Params.'' represents the number of parameters. FLOPs is computed based on input image sequence with a resolution of 20$\times$256$\times$256. $^*$ represents the final model. Best results are shown in boldface.}}}\label{tab-TSF}
	\begin{tabular}{|l|c|c|c|c|}
		\hline
		Frameworks&\#Params.$\downarrow$&FLOPs$\downarrow$&$P_d$$\uparrow$&$F_a$$\downarrow$\\\hline
		ResUNet w/o&1.017M&150.122G&85.243 &47.224 \\
		ResUNet-TM&1.017M&150.122G&87.697 &44.587 \\
		ResUNet-TSM&1.012M &150.895G&89.873 &16.201 \\
		ResUNet-TSFM$(*)$&1.012M &150.895G&91.579 &18.579 \\
		\hline
	\end{tabular}
\end{table}

Recurrent framework performs iterative feature refinement, and thus fully exploits the long-term temporal dependency from the entire input image sequence to improve the temporal feature representation capability. To demonstrate the effectiveness of our design choice, we introduce three kinds of variants to make comparisons with widely used temporal sliding window framework, and investigate the influence of input sequence length for training and test. We also present the results of ResUNet without temporal information exploitation as the baseline results. For fair comparison, the model sizes of all variants are set comparable, and the details of each variant are listed as follows:
\begin{itemize}
	\item \textbf{ResUNet-TSW$_t$}: We employ temporal sliding window framework of temporal length $t$ and stride 1 ($t$ image inputs for a single image output). $t$ is set to 5, 7, 20 in our experiments. For feature fusion, we first concatenate the input features along the channel dimension, which is then sent to several cascaded convolutional layers.
	\item \textbf{ResUNet-R$_{t-t}$}: We employ recurrent framework with temporal length $t$ for training. For test, we employ a temporal sliding window of temporal length $t$ and stride $t$ ($t$ image inputs for $t$ image outputs), and recurrent feature refinement is within each temporal sliding window. $t$ is set to 20 in our experiments.
	\item \textbf{ResUNet-R$_{t-seq}$}: We employ recurrent framework with temporal length $t$ for training, and the entire image sequence is used for test. $t$ is set to 10, 20, 40 in our experiments.
\end{itemize}

Table~\ref{tab-Different_Frameworks} shows the comparative results achieved by recurrent framework and its variants. It can be observed that improvements can be achieved among all variants against the baseline methods, which demonstrate the effectiveness of temporal additional information for IRST detection. All recurrent variants perform superior than temporal sliding window variants for 9.590 $P_d$ and 20.402 $F_a$ on average, which demonstrates the superiority of recurrent framework. Within the temporal sliding window variants, the detection performance increases first (\textit{i.e.,} 2.055 gain in $P_d$ and 5.681 drop in $F_a$ from 5 to 7) and then decreases as the number of input frames increases (\textit{i.e.,} 8.316 drop in $P_d$ from 7 to 20). This is because, appropriate temporal information can introduce performance improvement, while too much temporal information with inferior fusion methods deteriorates performance instead by ambiguous and misleading information. 

\begin{figure}[t]
	\centering\includegraphics[width=9cm]{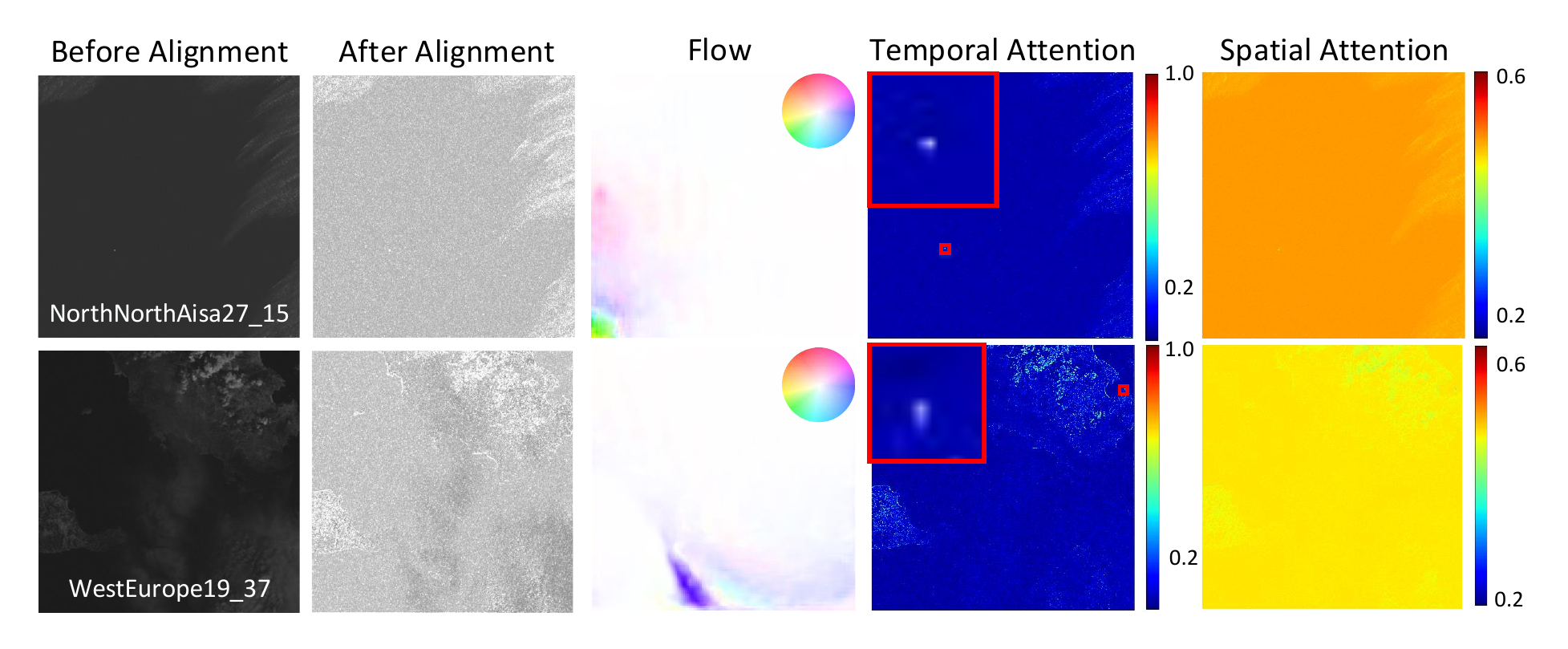}
	\vspace{-.3cm}
	\caption{Ablation on the temporal modulation (TM) and spatial modulation (SM) of TSFM. We show the temporal attention and spatial attention generated by TM and SM. Features before and after alignment, together with the flow derived by RAFT \cite{RAFT} are also presented. We also present the zoom-in target region for better visualization. Flow field color coding scheme is shown on the right. The direction and magnitude of the displacement vector are represented by hue and color intensity, respectively.}\label{fig-visual_attention}
	\vspace{-.2cm}
\end{figure}

For recurrent variants, due to limited GPU memory and batch-based training process, temporal length of the input video clip is manually pre-defined and fixed for training. During the test phase, the whole sequence (usually contains more than 200 images, and sometimes over 900 images) can be input into the network for recurrent feature refinement. Therefore, we investigate the influence of the input sequence length during the training and test phases. It can be observed from Table~\ref{tab-Different_Frameworks} that the performance improves steadily as the sequence length increases. This is because, recurrent frameworks effectively use the long-term temporal dependency for substantial performance improvement. Note that, as the number further increases (from 20 to 40), the improvement tends to be saturated. For a better trade-off of performance and efficiency, we employ the temporal length of 20 in the final training process. In addition, the $P_d$ and $F_a$ values tested on recurrent framework with fixed temporal sliding window are inferior to these on recurrent framework with entire sequence (85.726 vs. 86.468 in $P_d$ and 19.943 vs. 18.574 in $F_a$), which also demonstrate the effectiveness and superiority of recurrent framework.

\subsubsection{Pyramid Deformable Alignment}

Pyramid deformable alignment is used for coarse-to-fine motion compensation specific for satellite motion. To investigate the benefits introduced by this module, we compare our PDA (\textit{i.e.,} ResUNet-PDA) with four variants, and the details of each variant are listed as follows:

\begin{figure}[t]
	\centering\includegraphics[width=9cm]{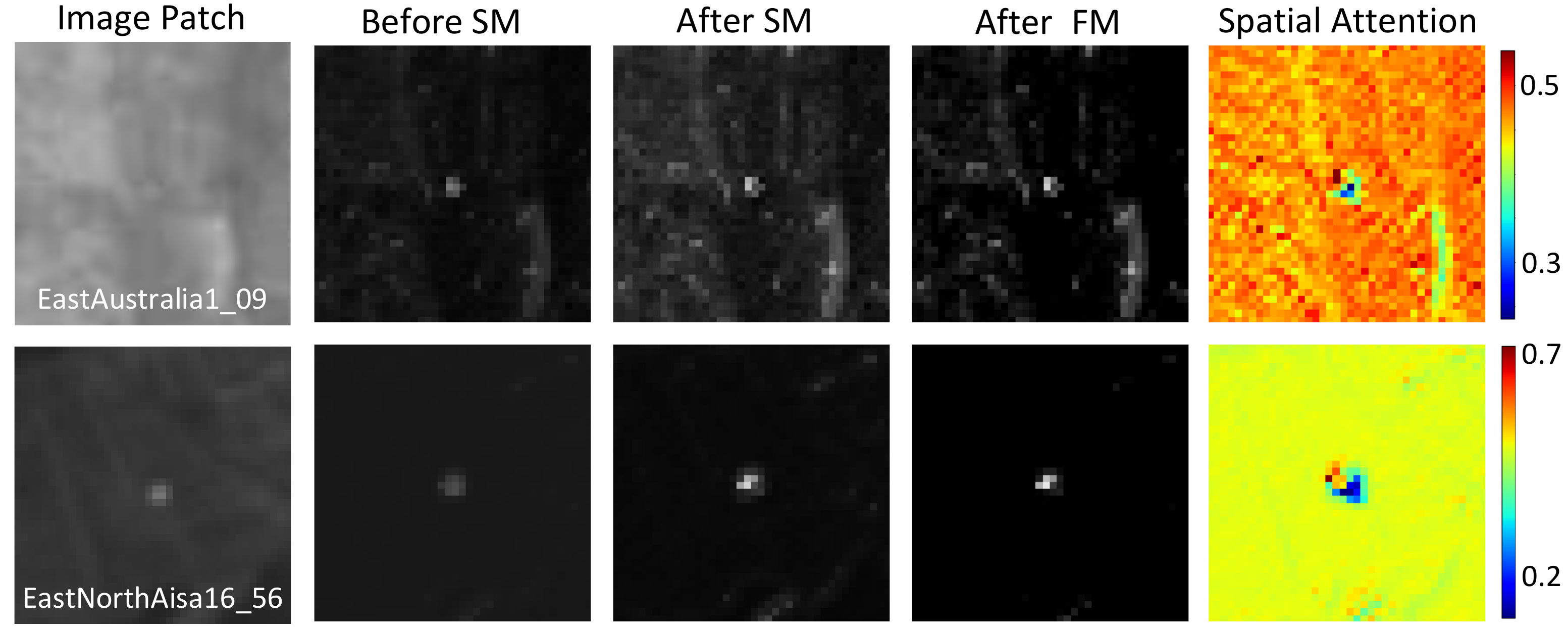}
	\vspace{-.3cm}
	\caption{Ablation on the spatial modulation (SM) and frequency modulation (FM) of TSFM. We show the zoom-in target regions of image, features before and after SM, FM. The zoom-in spatial attention is also presented.}\label{fig-visual_SF}
	\vspace{-.2cm}
\end{figure}

\begin{itemize}
	\item \textbf{ResUNet w/o}: We remove PDA to test its effectiveness for feature alignment in this variant. 
	\item \textbf{ResUNet-Flow}: We replace the deformable convolution layer in PDA by optical flow estimation using several cascaded convolution and subsequent warping for two-stage feature alignment in this variant.
	\item \textbf{ResUNet-Att}: We replace the deformable convolution layer in PDA by recent popular attention mechanism (\textit{i.e.,} cross attention in specific) for weighted summation-based feature alignment in this variant. 
	\item \textbf{ResUNet-DA}: We eliminate the pyramid structure in PDA by only a single deformable convolution layer for single-scale feature alignment in this variant.
\end{itemize}

Table~\ref{tab-PDA} shows the comparative results achieved by pyramid deformable alignment and its variants. It can be observed that deformable convolution for feature alignment is superior than optical flow-based method for 5.011 $P_d$ gain and 34.913 $F_a$ drop. This is because, deformable convolution employs implicit motion compensation in a unified step, avoiding ambiguous and duplicate results caused by wrapping. In addition, attention-based methods perform the worst, and even worse than flow-based methods. This is because, the complex background clutter and variable target appearance introduce much noise for spatial weighted summation operation of attention-based methods. In addition, pyramid structure can introduce 0.853 $P_d$ gain, which further demonstrates the effectiveness of coarse-to-fine feature fusion. In Fig.~\ref{fig-visual_align}, we show adjacent images, corresponding features, together with aligned features, and depict the flow (derived by RAFT \cite{RAFT}) between reference and aligned features. Compared with the flow without feature alignment, the flows of alignment modules are much smaller and cleaner. Among them, PDA achieves the smallest and cleanest flow results. It is demonstrated that PDA module can successfully compensate motions introduced by moving satellite. 

\subsubsection{Temporal-Spatial-Frequent Modulation}

Temporal-spatial-frequent modulation (TSFM) is used for dynamic feature aggregation and adaptive feature enhancement. To investigate the benefits introduced by this module, we compare our TSFM with four variants, and the details of each variant are listed as follows:

\begin{itemize}
	\item \textbf{ResUNet w/o}: We remove TSFM to test its effectiveness for temporal feature aggregation and spatial-frequency feature enhancement in this variant. 
	\item \textbf{ResUNet-TM}: We replace TSFM by temporal modulation using temporal attention for temporal feature aggregation in this variant.
	\item \textbf{ResUNet-TSM}: We replace TSFM by temporal-spatial modulation using cascaded temporal attention and spatial attention for temporal feature aggregation and spatial feature enhancement in this variant. 
\end{itemize}

\begin{table}[t]
	\footnotesize
	\setlength{\tabcolsep}{2.9mm}
	\renewcommand\arraystretch{1.2}
	\centering
	\caption{{{$P_d$($\times 10^2$) and $F_a$($\times 10^6$) values achieved by ResUNet\_RFR on 5 image sequences with one-pixel targets and 5 image sequences with sub-pixel targets.}}}\label{tab_pixel}
	\begin{tabular}{|cc|cc|cc|}
		\hline
		\multicolumn{2}{|c|}{One-pixel targets}&\multicolumn{2}{c|}{Sub-pixel targets}&	\multicolumn{2}{|c|}{Total}\\\hline
		$P_d$$\uparrow$&$F_a$$\downarrow$&$P_d$$\uparrow$&$F_a$$\downarrow$&$P_d$$\uparrow$&$F_a$$\downarrow$\\\hline
		86.03&30.80&82.52&18.77&84.37&24.95\\
		\hline
	\end{tabular}
\end{table}

\begin{figure}[t]
	\centering
	\includegraphics[width=9cm]{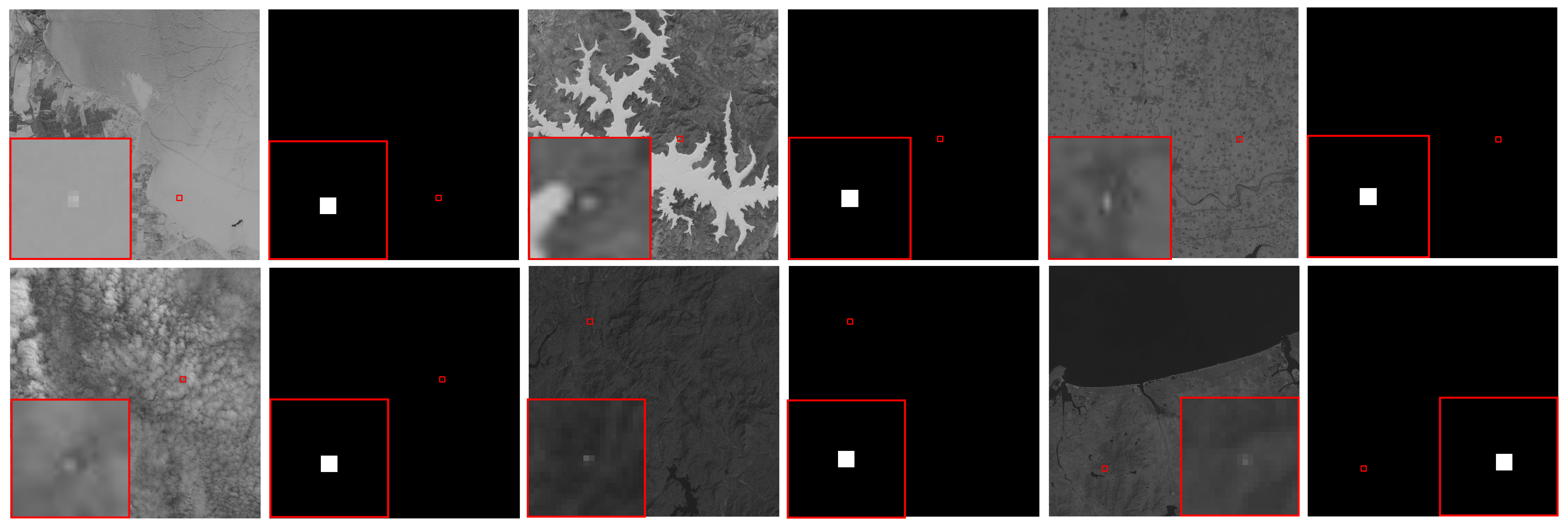}
	\vspace{-.5cm}
	\caption{{Qualitative results of ResUNet\_RFR method on one-pixel targets (row 1) and sub-pixel targets (row 2).}}\label{fig_subpixel}
	\vspace{-.2cm}
\end{figure}

Table~\ref{tab-TSF} shows the comparative results achieved by temporal-spatial-frequent modulation (TSFM) and its variants. It can be observed that temporal modulation introduces 2.454 $P_d$ gain and 2.637 $F_a$ drop compared with ResUNet-w/o, which demonstrates its effectiveness for feature fusion. In Fig.~\ref{fig-visual_attention}, we visualize the flow between the reference feature and aligned feature, together with the temporal attention map. It can be observed from the visualization of flow and temporal attention that neighborhood features can be accurately aligned, and most of the background clutters are suppressed while targets are correctly enhanced and propagated.

Spatial modulation (SM) and frequency modulation (FM) are used for feature enhancement. Ablation results in Table~\ref{tab-TSF} show substantial performance improvements introduced by SM and FM (2.176 and 1.706 gain for $P_d$). In addition, we visualize the spatial attention map, together with the features and corresponding zoom-in target region before and after SM and FM in Fig.~\ref{fig-visual_SF}. It can be observed that spatial attention and frequency attention effectively enhance the response of high-intensity and high-frequency regions, especially for small objects. 

\subsection{Discussion}\label{sec-Discussion}
In this subsection, we investigate the generalization of RFR on one-pixel targets and sub-pixel targets. Specifically, we synthesized 5 image sequences with one-pixel targets and 5 image sequences with sub-pixel targets. Based on the synthetic data, we employ ResUNet\_RFR as the baseline algorithm to conduct evaluation experiments. Quantitative and qualitative results in Table~\ref{tab_pixel} and Fig~\ref{fig_subpixel} show that ResUNet\_RFR can perform robust detection on one-pixel and sub-pixel targets.

\section{Conclusion}\label{sec-conclusion}
In this paper, we propose a recurrent feature refinement (RFR) framework to achieve MIRST detection in satellite videos. Different from existing sliding window-based methods, RFR recurrently performs effective and efficient feature alignment, propagation, aggregation and refinement by recurrent refinement block. Specifically, pyramid deformable alignment module integrates implicit motion compensation and object detection in an end-to-end manner. Temporal spatial frequency modulation module performs dynamic feature aggregation, and exploits the spatial and frequency saliency of IRST to preserve and enhance the target in recurrent feature propagation. Moreover, we develop an open-source MIRST dataset in satellite video to evaluate the performance of MIRST detection methods in satellite videos. Experimental results have shown the effectiveness and superiority of our method over the state-of-the-art methods.

\bibliographystyle{IEEEtran}
\bibliography{paper}

\vspace{-1.4cm}
\begin{IEEEbiography}
	[{\includegraphics[width=1in,height=1.25in,clip,keepaspectratio]{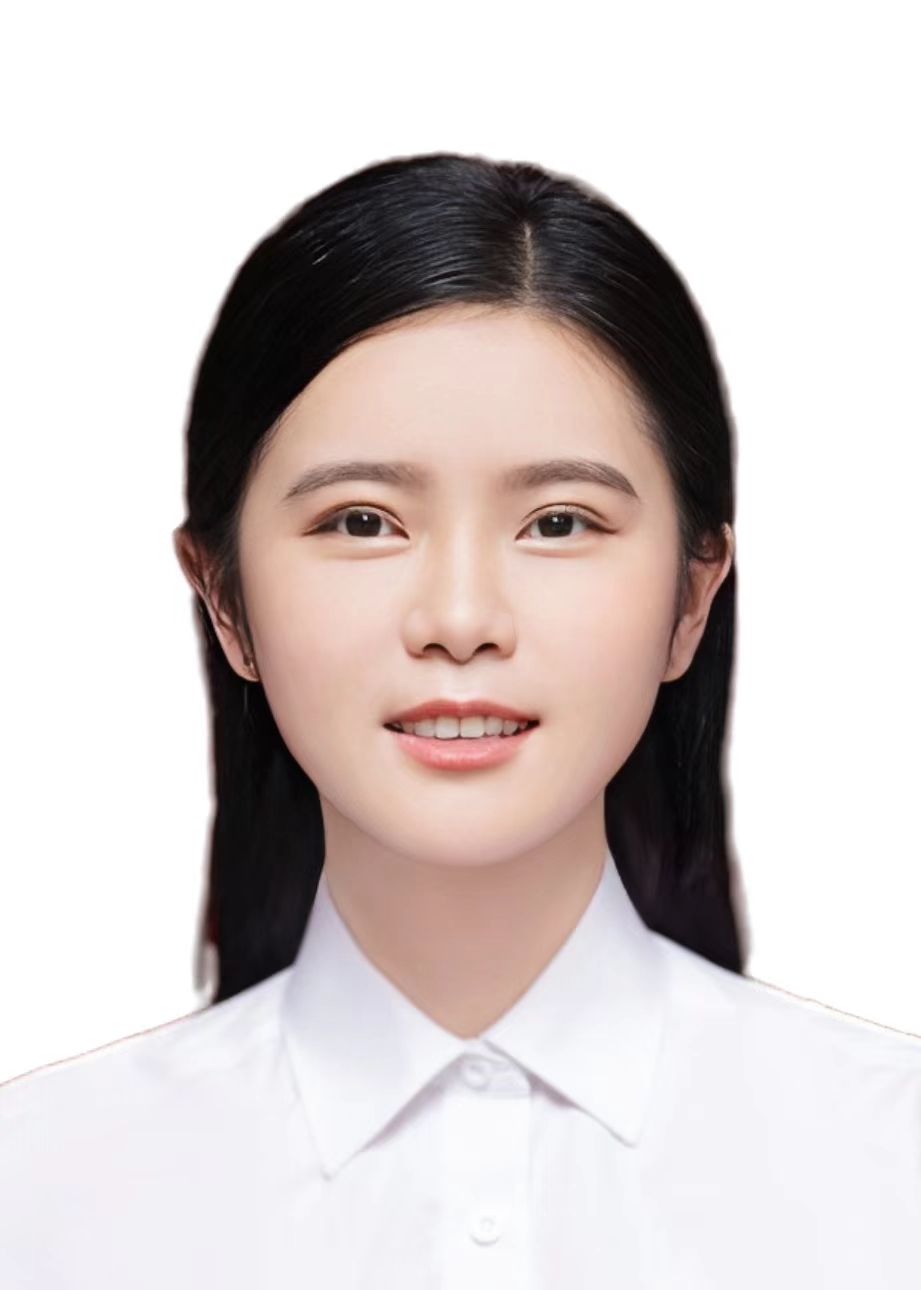}}]
	{Xinyi Ying} received the M.E. degree in information and communication engineering from National University of Defense Technology (NUDT), Changsha, China, in 2021. She is currently pursuing the Ph.D. degree with the College of Electronic Science and Technology, NUDT. Her research interests focus on detection and tracking of infrared small targets.
\end{IEEEbiography}

\vspace{-1.4cm}
\begin{IEEEbiography}
	[{\includegraphics[width=1in,height=1.25in,clip,keepaspectratio]{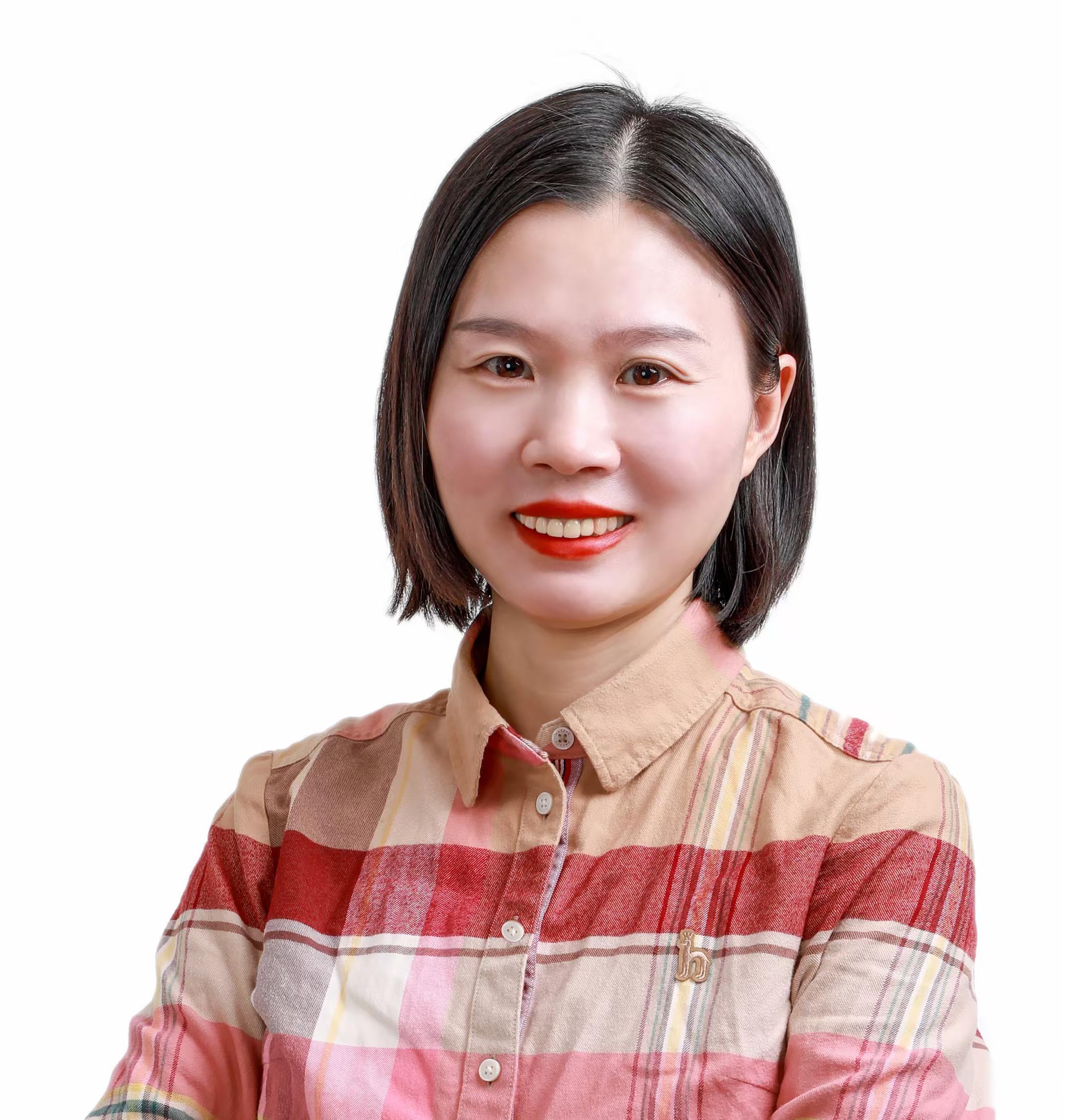}}]
	{Li Liu} received the Ph.D. degree in information and communication engineering from the National University of Defense Technology (NUDT), China, in 2012. From 2008 to 2010, she worked as a Visiting Student at the University of
	Waterloo, Canada, advised by Professor Paul Fieguth. From 2015 to 2016, she spent ten months visiting the Multimedia Laboratory at the Chinese University of Hong Kong, working with Professor Xiaogang Wang. From 2016.12 to 2018.9, she worked as a senior researcher at CMVS of the University of Oulu, Finland, working with Professor Matti Pietikainen. She was a cochair of nine International Workshops at CVPR, ICCV, and ECCV. She served as guest editor for two special
	issues in IEEE TPAMI and one in IJCV. She serves as an Associate Editor for IEEE Transactions on Circuits and Systems for Video Technology, IEEE Transactions on Geoscience and Remote Sensing and Pattern Recognition. She has been rated as one of the Most Cited Researchers in China. Her current research interests include Computer Vision, Machine Learning, Artificial Intelligence, Trustworthy AI, Synthetic Aperture Radar Imagery Analysis. Her papers have
	currently over 14,000 citations in Google Scholar (Until 2024.6.20). Her recent survey paper titled Deep Learning for Generic Object Detection.
\end{IEEEbiography}

\vspace{-1.4cm}
\begin{IEEEbiography}
	[{\includegraphics[width=1in,height=1.25in,clip,keepaspectratio]{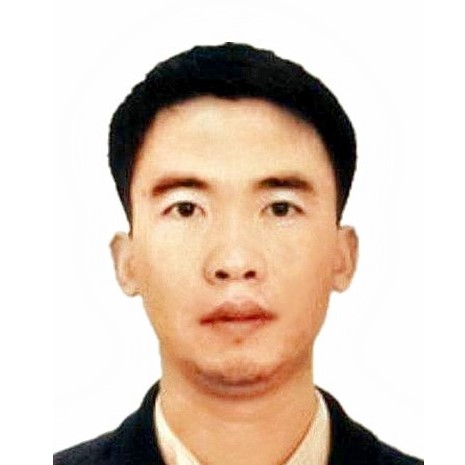}}]
	{Zaiping Lin} received the B.Eng. and Ph.D. degrees from the National University of Defense Technology (NUDT), Changsha, China, in 2007 and 2012, respectively. He is currently an Associate Professor with the College of Electronic Science and Technology, NUDT. His research interests include infrared image processing and signal processing
\end{IEEEbiography}

\vspace{-1.4cm}
\begin{IEEEbiography}
	[{\includegraphics[width=1in,height=1.25in,clip,keepaspectratio]{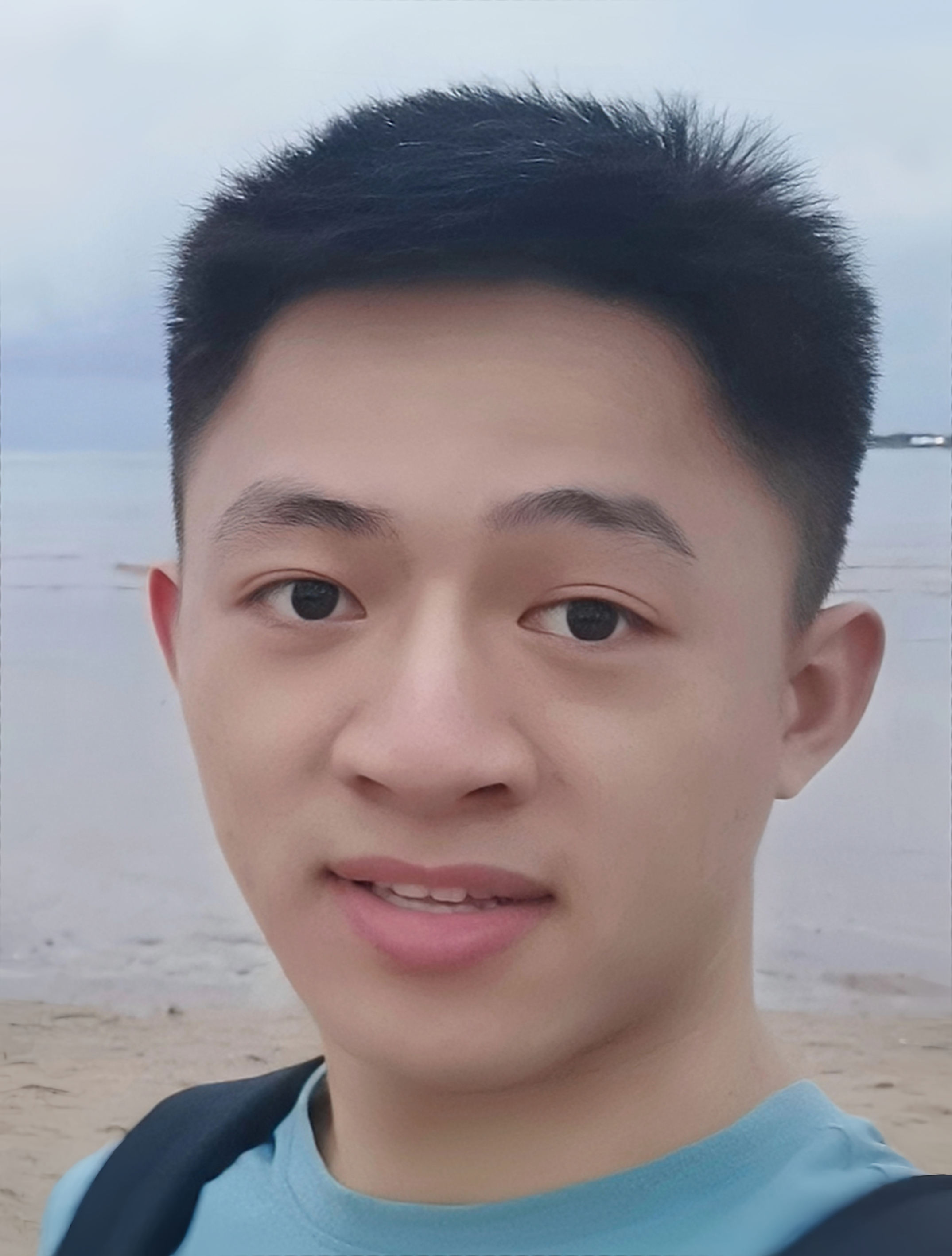}}]
	{Yangsi Shi} received his B.E. degree in Communication Engineering from Sun Yat-sen University, Shenzhen, China, in 2023. He is currently pursuing his M.E. degree in Information and Communication Engineering at the National University of Defense Technology (NUDT), Changsha, China. His research interests include target detection, with a particular focus on infrared small and weak target detection.
\end{IEEEbiography}

\begin{IEEEbiography}
	[{\includegraphics[width=1in,height=1.25in,clip,keepaspectratio]{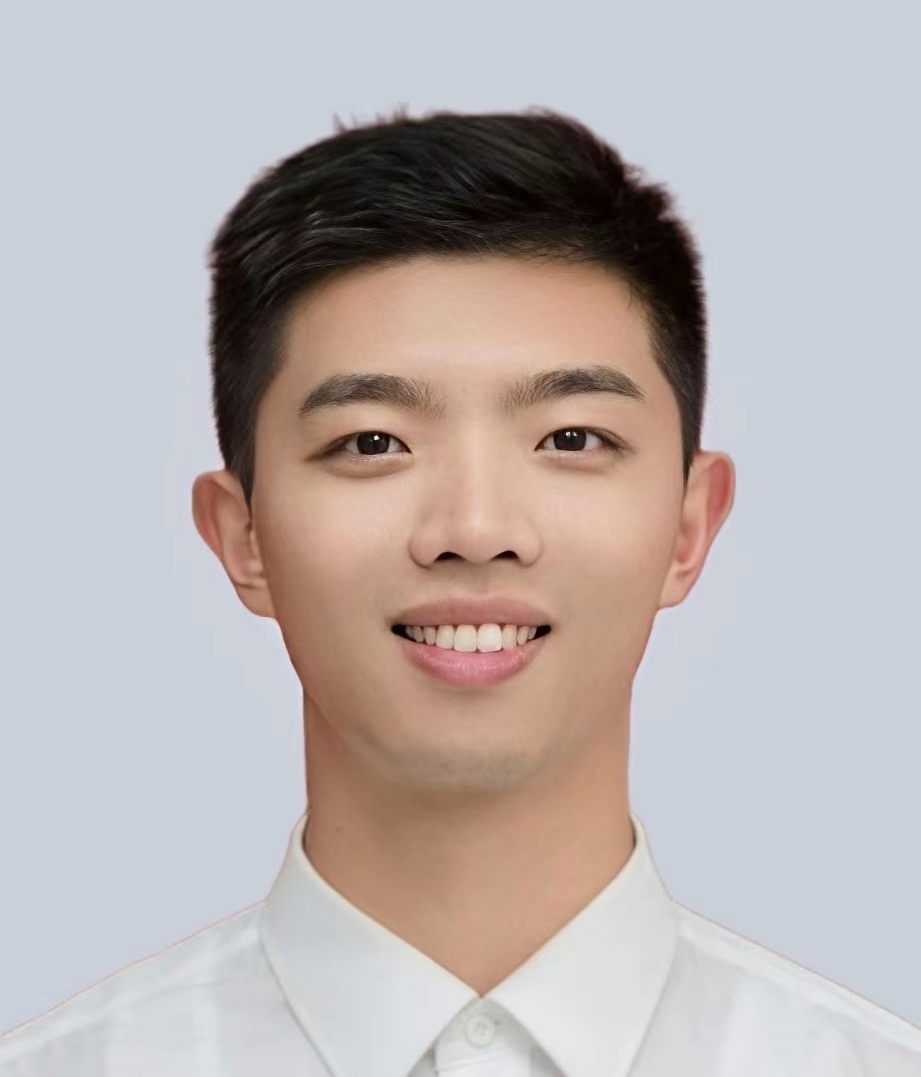}}]
	{Yingqian Wang} received his B.E. degree in electrical engineering from Shandong University, Jinan, China, in 2016, the Master and the Ph.D. degrees in information and communication engineering from National University of Defense Technology (NUDT), Changsha, China, in 2018 and 2023, respectively. Dr. Wang is currently an assistant professor with the College of Electronic Science and Technology, NUDT. His research interests focus on light field image processing, image super-resolution and infrared small target detection.
\end{IEEEbiography}

\begin{IEEEbiography}
	[{\includegraphics[width=1in,height=1.25in,clip,keepaspectratio]{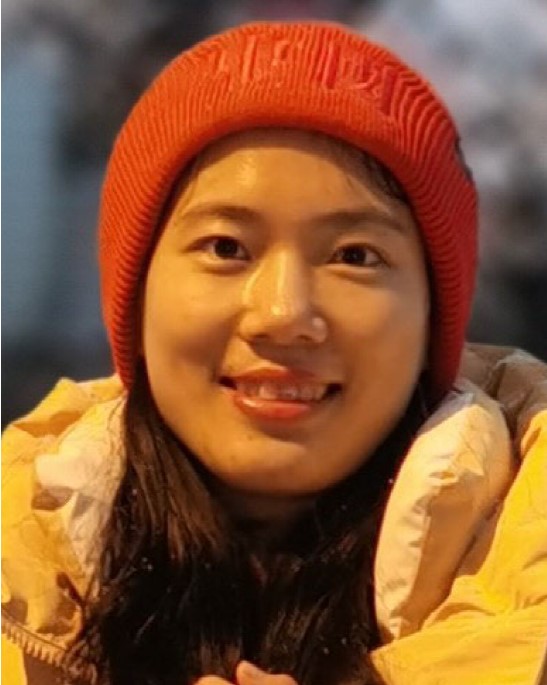}}]
	{Ruojing Li} received the B.E. degree in electronic engineering from the National University of Defense Technology (NUDT), Changsha, China, in 2020, where she is currently pursuing the Ph.D. degree in information and communication engineering. Her research interests include infrared small target detection, particularly on multiframe detection and deep learning.
\end{IEEEbiography}

\begin{IEEEbiography}
	[{\includegraphics[width=1in,height=1.25in,clip,keepaspectratio]{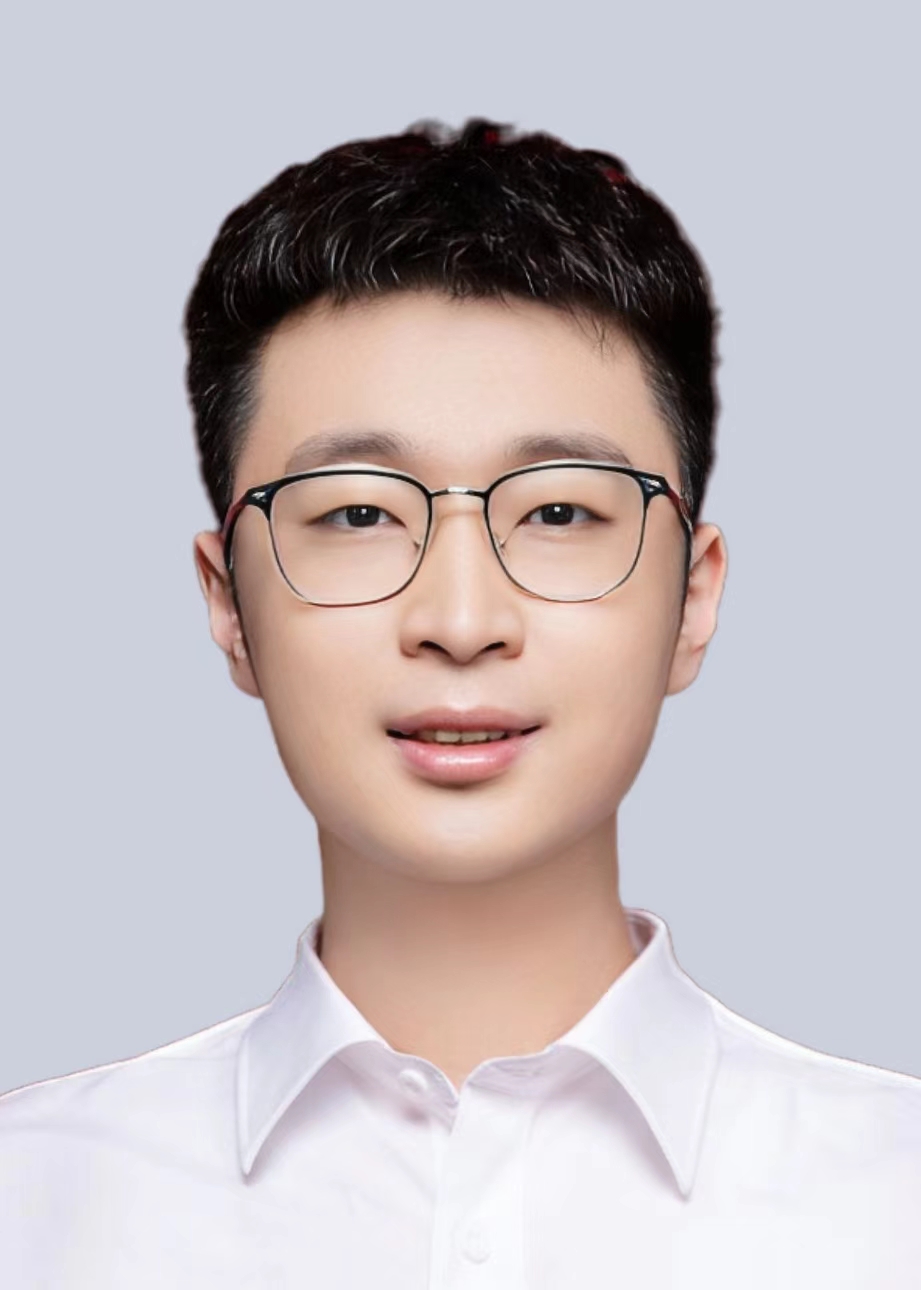}}]
	{Xu Cao} received the B.E. degree in information engineering in 2018, and the M.E. degree in information and communication engineering from National University of Defense Technology (NUDT), Changsha, China, in 2021. He is currently pursuing the Ph.D. degree with the College of Electronic Science and Technology, NUDT. His research interests focus on object detection and multi-source information fusion interpretation.
\end{IEEEbiography}

\begin{IEEEbiography}
	[{\includegraphics[width=1in,height=1.25in,clip,keepaspectratio]{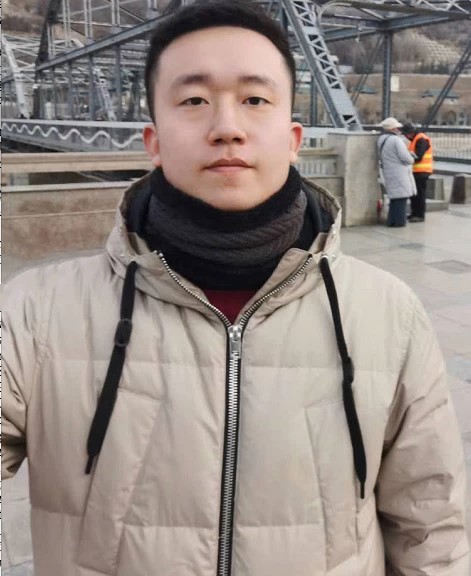}}]
	{Boyang Li} received the Ph.D and Master degrees from NUDT and National Defense Technology Innovation Institute in 2024 and 2020, respectively. Before that, He received the B.E. degree from Tianjin University in 2017. Currently, he is an assistant professor with the College of Electronic Science and Technology, NUDT. His research interests focus on optical image processing, interpretation and application, particularly on infrared small target detection, weakly supervised semantic segmentation, and neural network compression and acceleration.
\end{IEEEbiography}

\begin{IEEEbiography}
	[{\includegraphics[width=1in,height=1.25in,clip,keepaspectratio]{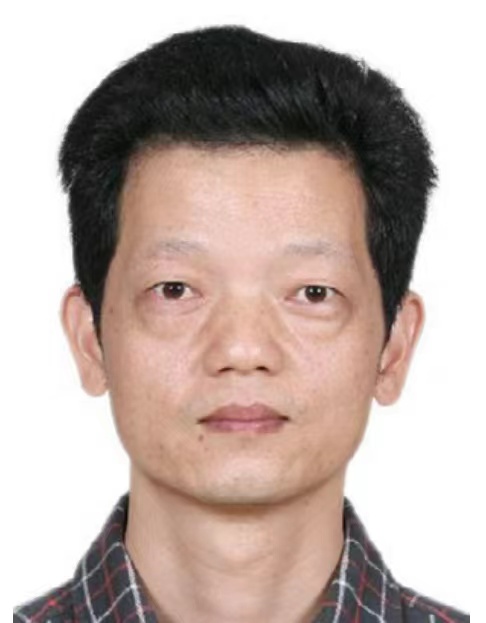}}]
	{Shilin Zhou} is currently a Professor with the College of Electronic Science and Technology, National University of Defense Technology, China. His main research interests include pattern recognition, signal processing, computer vision, intelligent information processing, and remote sensing image processing.
\end{IEEEbiography}

\begin{IEEEbiography}
	[{\includegraphics[width=1in,height=1.25in,clip,keepaspectratio]{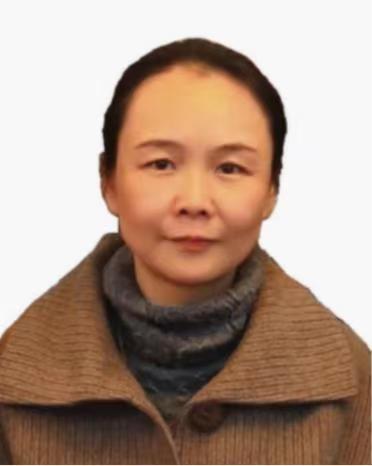}}]
	{Wei An} received the Ph.D. degree from the National University of Defense Technology (NUDT), Changsha, China, in 1999. She was a Senior Visiting Scholar with the University of Southampton, Southampton, U.K., in 2016. She is currently a Professor with the College of Electronic Science and Technology, NUDT. She has authored or co-authored over 100 journal and conference publications. Her current research interests include signal processing and image processing.
\end{IEEEbiography}

\end{document}